\documentclass[twoside,11pt]{article}

\usepackage{jmlr2e}

\usepackage{amsmath,amssymb,latexsym}
\usepackage{graphicx, color}
\usepackage{float,epsfig,subfig,epstopdf}
\usepackage{graphicx}

\usepackage{algorithm}
\usepackage{algorithmic}
\usepackage{subfig}

\newcommand{\Ind}{\mathbb{I}}
\usepackage{comment}

\newcommand{\specialcell}[2][c]{%
	\begin{tabular}[#1]{@{}c@{}}#2\end{tabular}}

\ShortHeadings{Efficient Learning and Decoding of the CT-HMM}{Yu-Ying Liu, et. al}
\firstpageno{1}

\begin{document}
\title{Efficient Learning and Decoding of the Continuous-Time Hidden Markov Model for Disease Progression Modeling}
\author{\name Yu-Ying Liu \email yuyingliu@gatech.edu \\
       \addr College of Computing \\ 
       Georgia Institute of Technology\\
       Atlanta, GA USA
       \AND
       \name  Alexander Moreno $^\dagger$ \email alexander.f.moreno@gmail.com \\
       \addr College of Computing \\ 
       Georgia Institute of Technology\\
       Atlanta, GA USA
       \AND
       \name Maxwell A. Xu \stepcounter{footnote}\thanks{Denotes equal contribution to this work} \email maxxu@gatech.edu \\
       \addr College of Computing \\ 
       Georgia Institute of Technology\\
       Atlanta, GA USA
      \AND
      \name Shuang Li \email shuangli@fas.harvard.edu \\
      \addr Department of Statistics \\ 
      Harvard University \\
      Boston, MA USA
      \AND
      \name Jena C. McDaniel \email jena.c.mcdaniel@ku.edu \\
      \addr Department of Speech-Language-Hearing \\ 
      Kansas University\\
      Lawrence, KS USA
      \AND
      \name Nancy C. Brady \email nbrady@ku.edu \\
      \addr Department of Speech-Language-Hearing \\ 
      Kansas University\\
      Lawrence, KS USA
      \AND
      \name Agata Rozga \email agata@gatech.edu \\
      \addr College of Computing \\ 
      Georgia Institute of Technology\\
      Atlanta, GA USA
       \AND
       \name Fuxin Li \email lif@eecs.oreganstate.edu \\
       \addr School of Electrical Engineering and Computer Science\\
       Oregan State University\\
       Corvallis, Oregon USA
       \AND
       \name Le Song \email lsong@cc.gatech.edu \\
       \addr College of Computing \\ 
       Georgia Institute of Technology\\
       Atlanta, GA USA
       \AND
       \name James M. Rehg \email rehg@cc.gatech.edu \\
       \addr College of Computing \\ 
       Georgia Institute of Technology\\
       Atlanta, GA USA
	   }
\editor{x}

\maketitle

\begin{abstract}
The Continuous-Time Hidden Markov Model (CT-HMM) is an attractive approach to modeling disease progression due to its ability to describe noisy observations arriving irregularly in time. However, the lack of an efficient parameter learning algorithm for CT-HMM restricts its use to very small models or requires unrealistic constraints on the state transitions. In this paper, we present the first complete characterization of efficient EM-based learning methods for CT-HMM models, as well as the first solution to decoding the optimal state transition sequence and the corresponding state dwelling time. We show that EM-based learning consists of two challenges: the estimation of posterior state probabilities and the computation of end-state conditioned statistics. We solve the first challenge by reformulating the estimation problem as an equivalent discrete time-inhomogeneous hidden Markov model. The second challenge is addressed by adapting three distinct approaches from the continuous time Markov chain (CTMC) literature to the CT-HMM domain. Additionally, we further improve the efficiency of the most efficient method by a factor of the number of states. Then, for decoding, we incorporate a state-of-the-art method from the (CTMC) literature, and extend the end-state conditioned optimal state sequence decoding to the CT-HMM case with the computation of the expected state dwelling time.  We demonstrate the use of CT-HMMs with more than 100 states to visualize and predict disease progression using a glaucoma dataset and an Alzheimer's disease dataset, and to decode and visualize the most probable state transition trajectory for individuals on the glaucoma dataset, which helps to identify progressing phenotypes in a comprehensive way. Finally, we apply the CT-HMM modeling and decoding strategy to investigate the progression of language acquisition and development.
\end{abstract}



\section{Introduction}

The goal of disease progression modeling is to learn a model for the temporal evolution of a disease from sequences of clinical measurements obtained from a longitudinal sample of patients. By distilling population data into a compact representation, disease progression models can yield insights into the disease process through the visualization and analysis of disease trajectories. For example, models can be used to explore the most likely path and the expected time a disease has taken in terms of stages of the disease. The models can also be used to predict the future course of disease in an individual, supporting the development of individualized treatment schedules and improved treatment efficiencies. Furthermore, progression models can support phenotyping of disease progression by grouping patients based on their progression trajectories using a provided similarity measurement between trajectories.

Hidden variable models are particularly attractive for modeling disease progression for three reasons: 1) they can support the abstraction of a disease state via the latent variables and thus are easily interpretable; 2) they can deal with noisy measurements effectively; and 3) any
prior knowledge or constraints on the temporal evolution of the latent states can be
captured by a model of the state dynamics. The interpretability of latent state models
is an attractive feature: since the latent states have a direct interpretation in the context of an experiment, the examination of latent state trajectories (following model
fitting) is a potentially-powerful tool for gaining insight into complex temporal patterns.
This is particularly important if the probability distributions obtained from
latent state modeling are to be used in subsequent analysis steps, such as adjusting
the tailoring variables in a health intervention.

A standard latent variable model for temporal data is the Hidden Markov Model (HMM). The Discrete Time HMM (DT-HMM) \cite{RabinerIEEE1989} is widely used in speech recognition, robotics, signal processing, and other domains. It assumes that measurement data arrives at a fixed, regular sampling rate, and associates each measurement sample with an instantiated hidden state variable. The DT-HMM is an effective model for a wide range of time series data, such as the outputs of accelerometers, gyroscopes, and photoplethysmographic sensors. However, the fixed sampling rate assumptions that underly the DT-HMM make it an inappropriate model choice for data that is distributed irregularly in time, such as clinical data.

A further disadvantage of using DT-HMMs to model event data is the fact that transitions in the hidden state are assumed to occur at the sample times. Since event data may be distributed sparsely in time, a more flexible model would allow hidden state transitions to occur between observations. One potential approach to using DT-HMMs with event data would be to set the sampling period fine enough to describe the desired state dynamics and then use a missing data model to address the fact that many sample times will not have an associated measurement. While this approach is frequently-used, it has several undesirable properties. First, the treatment of missing measurements can be both inefficient and inaccurate when the number of observations is sparse relative to the sampling rate, but if the discretization is too coarse, many transitions could be collapsed into a single one, concealing the actual continuous-time dynamics. Second, the sparsity of measurement can itself change over time. For example, during the early stages of a disease visits to a doctor may be infrequent, while in the late stages a patient may need frequent visits to monitor the condition.  The need to tradeoff between the temporal granularity at which state transitions can occur and the number of missing measurements which must be handled is a consequence of using a discrete time model to describe an inherently sparse, continuous-time measurement process.

A \emph{Continuous-Time} HMM (CT-HMM) is an HMM in which both the transitions between hidden states and the arrival of observations can occur at arbitrary (continuous) times (\cite{CoxBook1965, JacksonJSS2011}). It is therefore suitable for irregularly-sampled temporal data such as clinical measurements (\cite{BartolomeoBMC2011,LiuMICCAI2013,Wang2014kdd,Dempsey2019, Nagesh2019}). Unfortunately, the additional modeling flexibility provided by CT-HMM comes at the cost of a more complex inference procedure. 
In CT-HMMs, not only are the hidden states unobserved, but the \emph{transition times} at which the hidden states are changing are also unobserved. Moreover, multiple unobserved hidden state transitions can occur between two successive observations.  Because of this, not only is expectation maximization (EM) based learning more difficult, but decoding the optimal sequence of state transitions and the expected dwelling times, which can be done directly via the Viterbi algorithm in a DT-HMM, also becomes difficult compared to a DT-HMM.

A previous method addressed these challenges in CT-HMM learning by directly maximizing the data likelihood (\cite{JacksonJSS2011}), but this approach is limited to very small model sizes. A general EM framework for continuous-time dynamic Bayesian networks, of which CT-HMM is a special case, was introduced in~\cite{nodelman05}, but that work did not address the question of efficient learning. Work by \cite{Jane2013StatMed} showed how to do efficient learning for a latent CTMC, which is similar to a CT-HMM but has a different observation model. Prior work by \cite{Wang2014kdd} used a closed form estimator due to \cite{Metzner2007,MetznerJournal2007} which assumes that the transition rate matrix can be diagonalized through an eigendecomposition during learning. However, it is possible that the transition matrix is non-diagonalizable or close to non-diagonalizable during learning leading to unstable results.  Consequently, there is a need for efficient and more general CT-HMM learning methods that can scale to large state spaces (e.g. hundreds of states or more)(\cite{MurilloNIPS2011Complete}) and that are robust to pathological transition matrices during learning.  For the decoding problem in CT-HMM, to the best of our knowledge, there is no prior work addressing the decoding of the optimal state transition path and the corresponding dwelling time in CT-HMM. Previous work only addresses the decoding of the optimal states \textit{at} the observation times in a CT-HMM (Jackson 2011).  A method to decode the optimal sequence of state transitions and the expected dwelling times is also needed for understanding the hidden dynamics of a given longitudinal dataset.

This article describes a computational framework for CT-HMM learning which can efficiently handle a large number of states and is robust to pathological transition matrices within an EM framework, and also presents the first solution to decoding the most likely state transitions and the corresponding expected dwelling times.  We build on our prior work in (\cite{LiuNips15, liu2017learning}), but include a novel approach to decoding the most likely state transitions and the corresponding expected dwelling times. We are the first to present a complete solution to the decoding problem for CT-HMMs. Our learning method also improves upon the complexity of (\cite{LiuNips15}) by a factor of the number of states. We demonstrate the utility of our approach to disease progression modeling on three different datasets representing diverse applications. One of the datasets, the language development in children with autism, is a new dataset not previously modeled in our prior work.

For EM-based learning in CT-HMM, we begin in Section~\ref{sec:ctmc} by introducing the mathematical definition of the Continuous-Time Markov Chain (CTMC). In a CTMC, the states are directly observable and there is no measurement process.  For EM-based learning, it turns out that the key computations which are required for CT-HMM learning also arise in fitting CTMC models to data (\cite{Metzner2007,Hobolth2011,TataruBio2011}). Section~\ref{sec:cthmm} describes the addition of a measurement process that extends the CTMC model into a CT-HMM, and introduces the key equations that arise in CT-HMM parameter learning. Our approaches to the CT-HMM learning problem using EM are presented in Section~\ref{sec:em}. These approaches differ in the specific computational methods used in the E-step, and are complementary to each other for handling different properties of the transition matrix. In Section~\ref{sec:exp}, we describe the experimental results of CT-HMM learning using both simulation studies and real-world clinical datasets for glaucoma and Alzheimer's disease. Our results show that the learned model has excellent predictive power for the disease's future course. In addition, the visualization of the model can help gain insights into a disease's progression characteristics in a more comprehensive way.

For the CT-HMM decoding problem, in section \ref{sec:intro-decode} we introduce the decoding problem in CTMCs and CT-HMMs, and in section \ref{sec:decode_state_path}, we review the most relevant work in CTMC from Levin et al. (2012). In section \ref{sec:state-trajectory-ct-hmm}, we present our approach for CT-HMM decoding, where the problem is decomposed into three parts: a) finding the optimal sequence of states at observation times b) finding the optimal sequence of states between successive observations c) finding the expected dwell times.  For a), we use the variant of the Viterbi algorithm for CT-HMMs that we developed in the EM learning section, for b) we adopt the method of \cite{Levin2012} (which was initally developed for CTMCs and extended to DT-HMMs by \cite{Grinberg2015}) and adapt it to the CT-HMM case, and for c) we borrow a transition matrix structure from \cite{Hajiaghayi2014} and combine it with our method for calculating the expected dwelling times, which was also developed in the EM learning section.  In Section \ref{sec:exp-decode}, we evaluate our method using simulation data and then visualize the decoded trajectories on the glaucoma dataset as well as a dataset on early language development for children with autism spectrum disorder.  We then demonstrate how the decoded continuous transition path in the state space can help in identifying the phenotypes of disease progression by direct visualization or through trajectory clustering methods. 

Our efficient learning and decoding methods for CT-HMM enable the use of CT-HMM for multivariate longitudinal clinical data with a large state space. The visualization of the learned model, prediction of the most probable future path, and decoding of the optimal state transition trajectory for a given individual data all support the understanding of the otherwise obscured hidden dynamics, which demonstrate the practical utility of CT-HMM for clinical data modeling. In particular, when applied to the language development use case, we are able to identify particular latent stalling states that serve as the initial barrier to full language development which hold clinical relevance in designing targeted intervention strategies. Note that additional updates will be made available through our project repository .~\footnote{https://github.com/rehg-lab/ct-hmm} 

\section{Learning in Continuous-time Hidden Markov Models}
\subsection{Continuous-Time Markov Chain}
\label{sec:ctmc}

\setlength{\abovedisplayskip}{5pt}
\setlength{\abovedisplayshortskip}{2pt}
\setlength{\belowdisplayskip}{5pt}
\setlength{\belowdisplayshortskip}{2pt}
\setlength{\jot}{4pt}
	
A continuous-time Markov chain (CTMC) is defined by a finite and discrete state space $S$, a state transition rate matrix $Q$, and an initial state probability distribution $\pi$. The elements $q_{ij}$ in $Q$ describe the rate at which the process transitions from state $i$ to $j$ for $i \neq j$, and $q_{ii}$ are specified such that each row of $Q$ sums to zero ($q_i = \sum_{j \neq i}{q_{ij}}$, $q_{ii} = -q_i$) (\cite{CoxBook1965}). 
In a time-homogeneous process, in which the $q_{ij}$ are independent of $t$, the sojourn time in each state $i$ is exponentially-distributed with parameter $q_i$: $f_i(t) = q_i e^{-q_i t}$ with mean $1/q_{i}$. 
The probability that the process's next move
is from state $i$ to state $j$ is given by $q_{ij}/q_i$.
If a realization of the CTMC is \emph{fully} observed, it means that one can observe every state transition time $(t'_0,t'_1,\ldots,t'_{V'})$, and the corresponding states $Y' = \{ y_0 = s(t'_0), ..., y_{V'} = s(t'_{V'}) \}$, where $s(t)$ denotes the state at time $t$.  In that case, the complete likelihood (CL) of the data is
\begin{align}
  \label{equ:CL}
  CL = \prod_{v'=0}^{V'-1} (q_{y_{v'},y_{v'+1}}/q_{y_{v'}}) (q_{y_{v'}} e^{-q_{y_{v'}} \tau_{v'}})
  = \prod_{v'=0}^{V'-1} q_{y_{v'},y_{v'+1}} e^{-q_{y_{v'}} \tau_{v'}} = \prod_{i=1}^{|S|}\left[\prod_{j=1,j\neq i}^{|S|} q_{ij}^{n_{ij}}\right] e^{-q_i\tau_i}  
\end{align}
where $\tau_{v'} = t'_{v'+1} - t'_{v'}$ is the time interval between two transitions, $n_{ij}$ is the number of transitions from state $i$ to $j$, and $\tau_i$ is the total amount of time the chain remains in state $i$.  
\begin{figure}[tb]
	\centering
	\includegraphics[width=10cm]{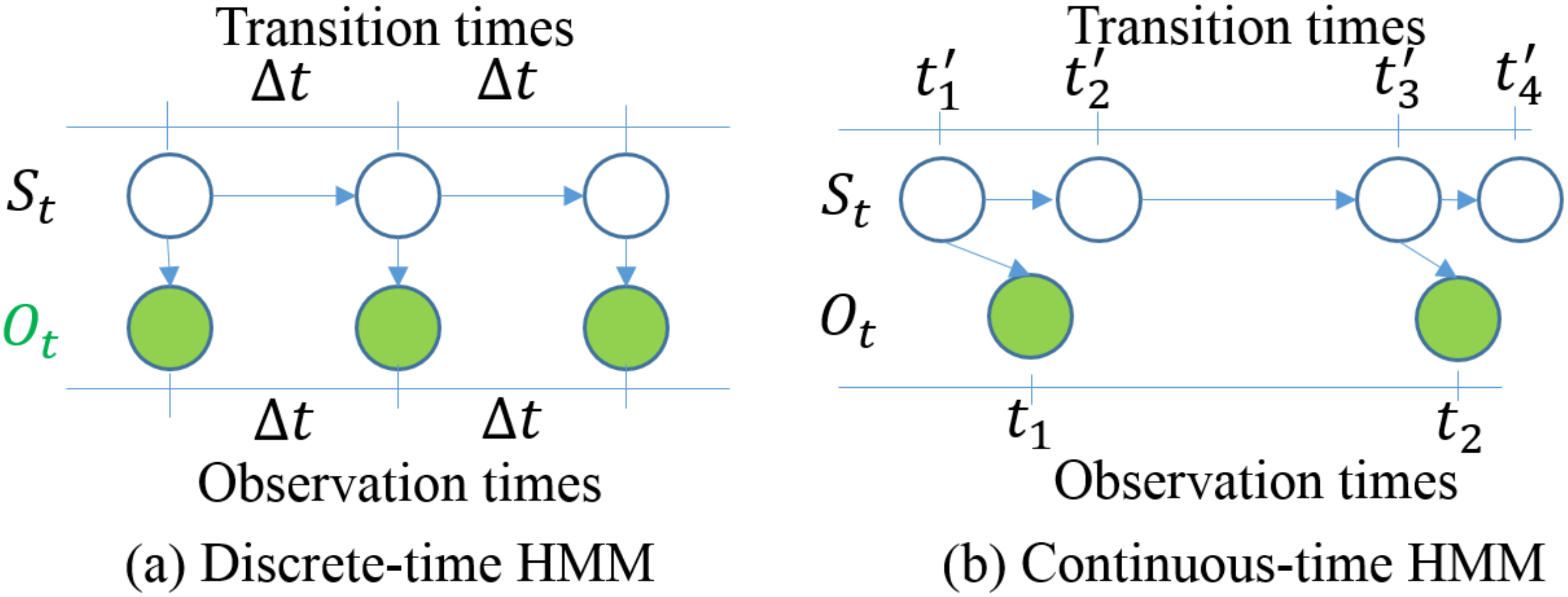}
	\vskip -0.05in
	\caption{	\small
		The DT-HMM and the CT-HMM.  In the DT-HMM, the observations $O_t$ and state transitions $S_t$ occur at fixed time intervals $\Delta_t$, and the states $S_t$ are the only source of latent information.  In the CT-HMM, the observations $O_t$ arrive at irregular time intervals, and there are two sources of latent information: the states $S_t$ and the transition times $(t_1',t_2', ...)$ between the states.}
	\label{fig:ct-hmm-vs-dt-hmm}
	\vskip -0.15in
\end{figure}

In general, a realization of the CTMC is observed only at \emph{discrete and irregular} time points $(t_0, t_1, ..., t_V)$, corresponding to a state sequence $Y$, which are \emph{distinct} from the transition times. As a result, the Markov process between two consecutive observations is \emph{hidden}, with potentially many unobserved state transitions. Thus, both $n_{ij}$ and $\tau_i$ are unobserved. To express the likelihood of the incomplete observations, we can utilize a discrete time hidden Markov model by defining a  state transition probability matrix for each time interval $t$, 
$P(t) = e^{Qt}$,
where $P_{ij}(t)$, the entry $(i,j)$ in $P(t)$, is the probability that the process is in state $j$ after time $t$, given that it is in state $i$ at time $0$. This quantity takes into account all possible intermediate state transitions and timing between $i$ and $j$ which are not observed. Then the likelihood of the data is
\begin{align}
\label{equ:L}
L =   \prod_{v=0}^{V-1} P_{y_v,y_{v+1}}(\tau_v)	  
   =   \prod_{v=0}^{V-1} \prod_{i,j=1}^{|S|} P_{ij}(\tau_v)^{\Ind(y_v = i, y_{v+1} = j)}
\end{align}
where $\tau_{v} = t_{v+1} - t_{v}$ is the time interval between two observations, $\Ind(\cdot,\cdot)$ is the indicator function that is $1$ if both arguments are true, otherwise it is 0. Note that there is no analytic maximizer of $L$, due to the structure of the matrix exponential, and direct numerical maximization with respect to $Q$ is computationally challenging. This motivates the use of an EM-based approach.

An EM algorithm for CTMC learning is described in~\cite{Metzner2007}. Based on Eq.~\ref{equ:CL}, the expected complete data log-likelihood takes the form
\begin{align}
\lbrace\sum_{i=1}^{|S|} \left[\sum_{j=1, j \neq i}^{|S|} \log(q_{ij}) \mathbb{E}[n_{ij} | Y, \hat{Q}_0]\right] - q_{i} \mathbb{E}[\tau_i | Y, \hat{Q}_0] \rbrace
\end{align}
where $\hat{Q}_0$ is the current estimate for $Q$, and $\mathbb{E}[n_{ij} | Y, \hat{Q}_0]$ and $\mathbb{E}[\tau_i | Y, \hat{Q}_0]$ are the expected state transition count and total duration given the incomplete observation $Y$ and the current transition rate matrix $\hat{Q}_0$, respectively. Once these two expectations are computed in the E-step, the updated $\hat{Q}$ parameters can be obtained via the M-step as 

\begin{align}
\hat{q}_{ij} &= \frac{\mathbb{E}[n_{ij} | Y, \hat{Q}_0]}{\mathbb{E}[\tau_i | Y, \hat{Q}_0]}, i \neq j
~~~ \text{and} ~~~ \hat{q}_{ii} = - \sum_{j \neq i} \hat{q}_{ij}. 
\end{align}

Now the main computational challenge is to evaluate $\mathbb{E}[n_{ij} | Y, \hat{Q}_0]$ and $\mathbb{E}[\tau_i | Y, \hat{Q}_0]$. By exploiting the properties of the Markov process, the two expectations can be decomposed as~(\cite{Bladt2005}):
{\footnotesize
\begin{align*}
  & \mathbb{E}[n_{ij} | Y, \hat{Q}_0] =
  \sum_{v=0}^{V-1} \mathbb{E} [n_{ij} | y_v, y_{v+1}, \hat{Q}_0] =  
  \sum_{v=0}^{V-1} \sum_{k,l=1}^{|S|} \Ind(y_v = k, y_{v+1} = l) \mathbb{E} [n_{ij} | y_v = k, y_{v+1} = l, \hat{Q}_0] \\
  & \mathbb{E}[\tau_i | Y, \hat{Q}_0] =
  \sum_{v=0}^{V-1} \mathbb{E} [\tau_i | y_v, y_{v+1}, \hat{Q}_0] =  
   \sum_{v=0}^{V-1} \sum_{k,l=1}^{|S|} 
  \Ind(y_v = k, y_{v+1} = l) \mathbb{E} [\tau_i | y_v = k, y_{v+1} = l,\hat{Q}_0 ]
\end{align*}
}
Thus, the computation reduces to computing the end-state conditioned expectations $\mathbb{E} [n_{ij} | y_v = k, y_{v+1} = l,\hat{Q}_0]$ and $\mathbb{E} [\tau_i | y_v = k, y_{v+1} = l,\hat{Q}_0]$, for all $k,l,i,j \in S$. These expectations are also a key step in CT-HMM learning, and Section~\ref{sec:em} presents our approach to computing them.  

\subsection{Continuous-Time Hidden Markov Model}  
\label{sec:cthmm}

In this section, we describe the continuous-time hidden Markov model (CT-HMM) for disease progression and our approach to CT-HMM learning. 

\subsubsection{Model Description}

In contrast to CTMC, where the states are directly observed, none of the states are directly observed in CT-HMM. Instead, the available observational data $o$ depends on the hidden states $s$ via the measurement model $p(o|s)$. In contrast to a conventional HMM, the observations $(o_0, o_1,\ldots, o_V)$ are only available at irregularly-distributed continuous points in time $(t_0, t_1,\ldots,t_V)$. As a consequence, there are two levels of hidden information in a CT-HMM. First, at observation time, the state of the Markov chain is hidden and can only be inferred from measurements. Second, the state transitions in the Markov chain between two consecutive observations are also hidden. As a result, a Markov chain may visit multiple hidden states before reaching a state that emits a noisy observation. This additional complexity makes CT-HMM a more effective model for event data in comparison to HMM and CTMC. But as a consequence the parameter learning problem is more challenging.  We believe we are the first to present a comprehensive and systematic treatment of efficient EM algorithms to address these challenges.

A \emph{fully observed} CT-HMM contains four sequences of information: the underlying state transition time $(t'_0,t'_1,\ldots,t'_{V'})$, the corresponding state $Y' = \{ y_0 = s(t'_0), ..., y_{V'} = s(t'_{V'}) \}$ of the hidden Markov chain, and the observed data $O=(o_0, o_1,\ldots, o_{V})$ at time $T=(t_0, t_1,\ldots,t_{V})$. Their joint complete likelihood can be written as
\begin{align}
  CL = \prod_{v'=0}^{V'-1} q_{y_{v'},y_{v'+1}} e^{-q_{y_{v'}} \tau_{v'}} \prod_{v=0}^{V} p(o_{v} | s(t_{v})) = \prod_{i=1}^{|S|}\left[\prod_{j=1,j\neq i}^{|S|} q_{ij}^{n_{ij}}\right] e^{-q_i\tau_i} \prod_{v=0}^{V} p(o_{v} | s(t_{v}))
\end{align}

We make two simplifying assumptions.  First, we assume that the observation time is independent of the states and the state transition times.  Second, we assume that individual state trajectories are homogeneous, in that all sequences share the same global rate and emission parameters, which do not vary over time.  With the first assumption, we do not require any further assumptions on the distribution of observation times.  Furthermore, the observation time is not informative of the state.

We will focus our development on the estimation of the transition rate matrix $Q$. Estimates for the parameters of the emission model $p(o|s)$ and the initial state distribution $\pi$ can be obtained from the standard discrete time HMM formulation~(\cite{RabinerIEEE1989}), but with time-inhomogeneous transition probabilities, which we describe in section \ref{sec:computing-posterior-state}.  That is, the transition rates stay constant, but in the discrete-time formulation, the transition probabilities vary over time.

\subsubsection{Parameter Estimation}

We now describe an EM-based method for estimating $Q$ from data.  Given a current parameter estimate $\hat Q_0$, the expected complete log-likelihood takes the form
\begin{align}
L(Q) &= \lbrace\sum_{i=1}^{|S|} \left[\sum_{j=1, j \neq i}^{|S|}  \log(q_{ij}) \mathbb{E}[n_{ij} | O, T, \hat{Q}_0]\right] - q_{i} \mathbb{E}[\tau_i | O, T, \hat{Q}_0] \rbrace \nonumber\\
&\qquad+\sum_{v=0}^V \mathbb{E}[\log p(o_v|s(t_v))|O,T,\hat Q_0] 
\end{align}

In the M-step, taking the derivative of $L$ with respect to $q_{ij}$ yields
\begin{align}
\hat{q}_{ij} &= \frac{\mathbb{E}[n_{ij} | O, T, \hat{Q}_0]}{\mathbb{E}[\tau_i | O, T, \hat{Q}_0]}, i \neq j
~~~ \text{and} ~~~ \hat{q}_{ii} = - \sum_{j \neq i} \hat{q}_{ij}.
\end{align}

The challenge lies in the E-step, where we compute the expectations of $n_{ij}$ and $\tau_i$ conditioned on the observation sequence.   The expectation for $n_{ij}$ can be expressed in terms of the expectations between successive pairs of observations as follows:
{\footnotesize
\begin{align}
E[n_{ij}|O,T,\hat{Q}_{0}]&=\sum_{s(t_{1}),...,s(t_{V})} p(s(t_{1}),...,s(t_{V})|O,T,\hat{Q}_{0})\mathbb{E}[n_{ij}|s(t_{1}),...,s(t_{V}),\hat{Q}_{0}]\\
&=\sum_{s(t_{1}),...,s(t_{V})} p(s(t_{1}),...,s(t_{V})|O,T,\hat{Q}_{0})\sum_{v=1}^{V-1}\mathbb{E}[n_{ij}|s(t_{v}),s(t_{v+1}),\hat{Q}_{0}]\\
&=\sum_{s(t_{1}),...,s(t_{V})}\sum_{v=1}^{V-1}p(s(t_{1}),...,s(t_{V})|O,T,\hat{Q}_{0})\mathbb{E}[n_{ij}|s(t_{v}),s(t_{v+1}),\hat{Q}_{0}]\\
&=\sum_{v=1}^{V-1}\sum_{s(t_{v}),s(t_{v+1})}p(s(t_{v}),s(t_{v+1})|O,T,\hat{Q}_{0})\mathbb{E}[n_{ij}|s(t_{v}),s(t_{v+1}),\hat{Q}_{0}]\nonumber \\
&\text{ by marginalization}\\
&= \sum_{v=1}^{V-1}\sum_{k,l=1}^{|S|}p(s(t_{v})=k,s(t_{v+1})=l|O,T,\hat{Q}_{0})\mathbb{E}[n_{ij}|s(t_{v})=k,s(t_{v+1})=l,\hat{Q}_{0}]\label{eq:endn}
\end{align}
}
In a similar way, we can obtain an expression for the expectation of $\tau_i$:
{\footnotesize
\begin{align}
  \mathbb{E}[\tau_i | O, T, \hat{Q}_0] = \sum_{v=1}^{V-1}
  \sum_{k,l=1}^{|S|}   p(s(t_v)=k, s(t_{v+1})=l | O, T, \hat{Q}_0) \mathbb{E} [\tau_{i} | s(t_v)= k, s(t_{v+1}) = l, \hat{Q}_0].\label{eq:endt}
\end{align}  
}
Note that, in contrast to the CTMC case, during CT-HMM learning we cannot observe the states directly at the observation times. Therefore, while the sum of expectations is weighted via indicator variables in the CTMC case, the weights are probabilities obtained through inference in the CT-HMM case. 

\begin{figure}[tb]
	\centering
	\includegraphics[height=4cm]{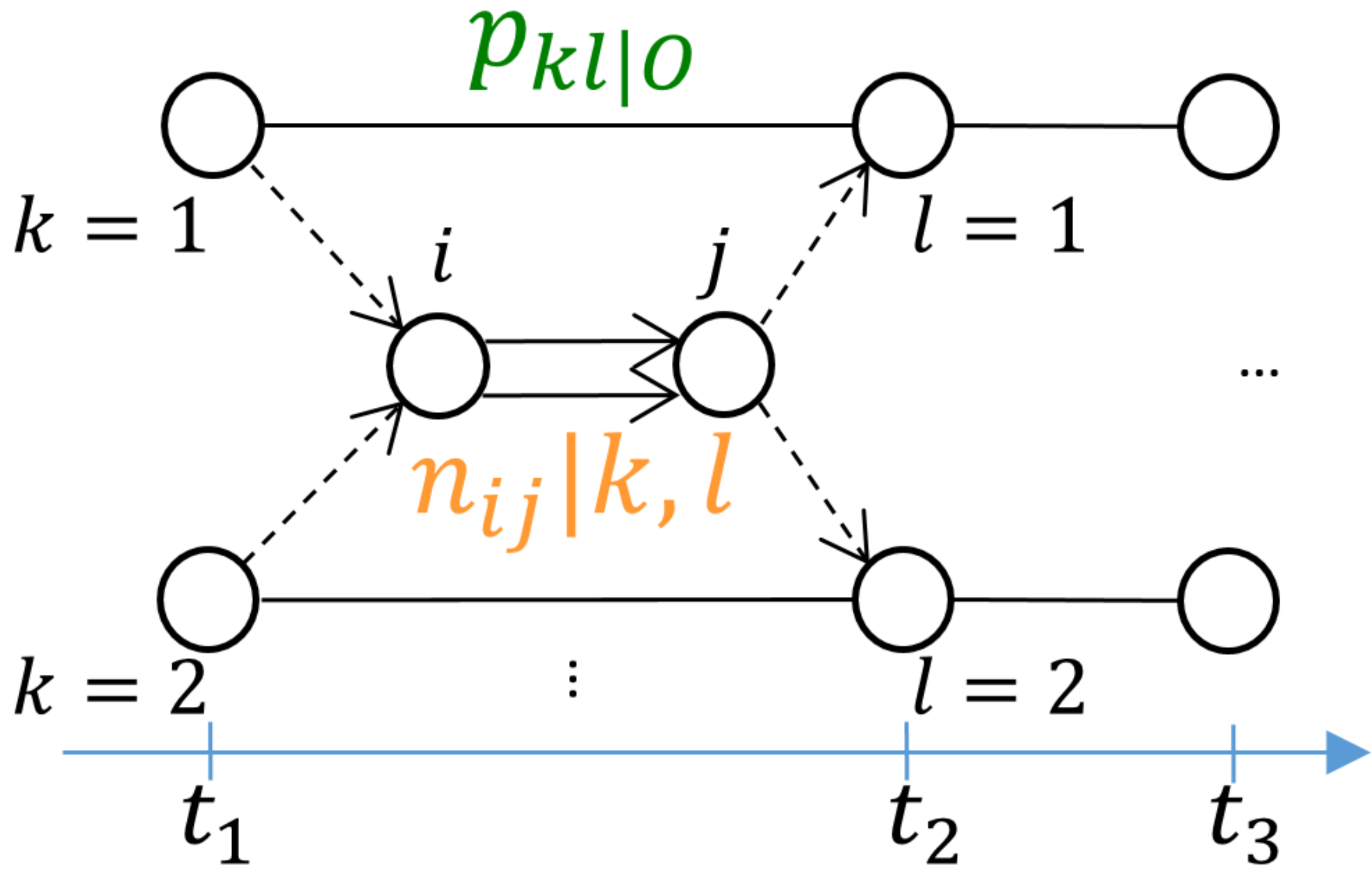}
        \caption{Illustration of the decomposition of the expectation calculations (equation \ref{eq:endn}) according to their inner-outer structure, where $k$ and $l$ represent the two possible end-states at successive observation times $(t_1, t_2)$, and $i,j$ denotes a state transition from $i$ to $j$ within the time interval. $p_{kl|O}$ represents $p(s(t_v) = k; s(t_{v+1}) = l | O, T,\hat{Q}_0)$ and $n_{ij}|k,l$ denotes $E[n_{ij}| s(t_v) = k, s(t_{v+1}) = l ,\hat{Q}_0]$ in equation \ref{eq:endn}.}
        \label{fig:inner-outer}
\end{figure}

The key to efficient computation of the expectations in Eqs.~\ref{eq:endn} and~\ref{eq:endt} is to exploit the structure of the summations. These summations have an inner-outer structure, which is illustrated in Fig.~\ref{fig:inner-outer}. The key observation is that the measurements partition the continuous timeline into intervals. It is therefore sufficient to compute the distribution over the hidden states at two successive observations, denoted by $p(s(t_{v})=k,s(t_{v+1})=l|O,T,\hat{Q}_{0})$, and use these probabilities to weight the expectation over unobserved state transitions, which we refer to as the \emph{end-state conditioned expectations} $\mathbb{E}[n_{ij}|s(t_{v}) = k,s(t_{v+1})=l,\hat{Q}_{0}]$ and $\mathbb{E} [\tau_{i} | s(t_v)= k, s(t_{v+1}) = l, \hat{Q}_0]$. We present three methods that can be used to compute the end-state conditioned expectations in Sec.~\ref{sec:em}. We now describe our approach to computing the hidden state probabilities at the observations.


\subsubsection{Computing the Posterior State Probabilities}
\label{sec:computing-posterior-state}
The challenge in efficiently computing $p(s(t_v)=k, s(t_{v+1})=l | O, T, \hat{Q}_0)$ is to avoid the explicit enumeration of all possible state transition sequences and the varying time intervals between intermediate state transitions (from $k$ to $l$).

\begin{figure}[t]
	\centering
	\includegraphics[height=4cm]{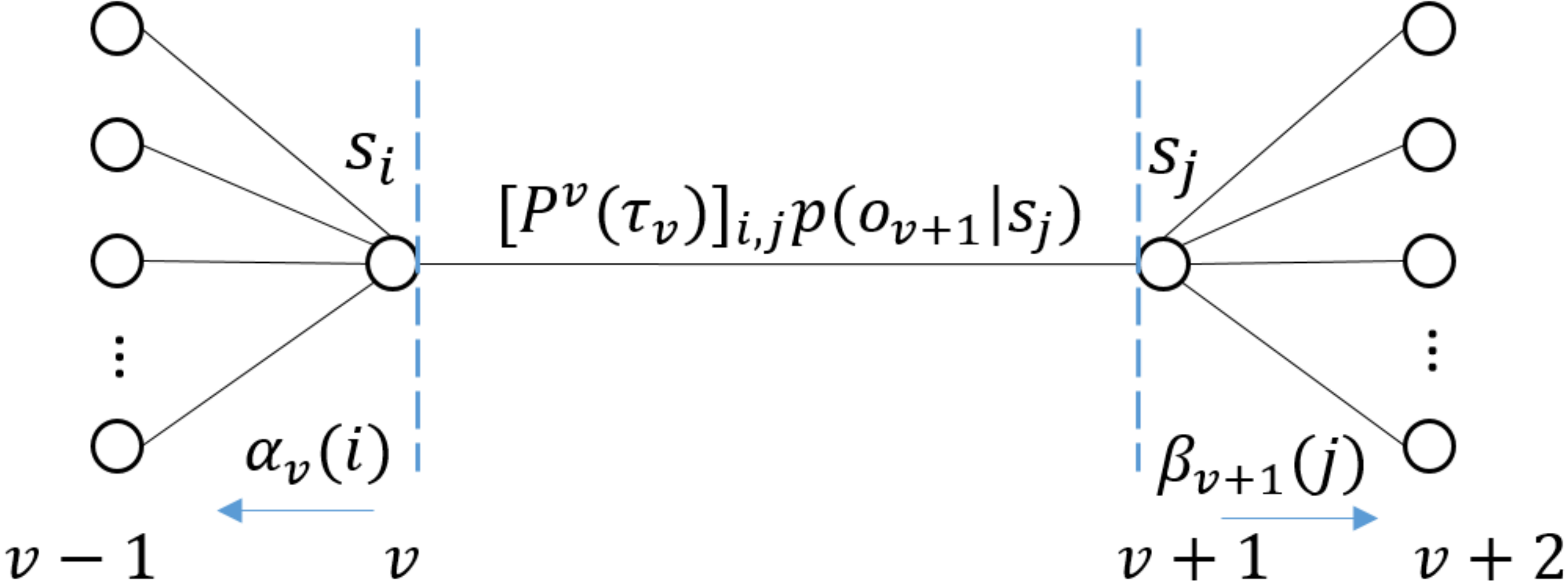}
        \caption{Illustration of the computation of the posterior state probabilities $p(s(t_v) = k, s(t_{v+1}) = l| O, T,\hat{Q}_0)$. An equivalent time-inhomogeneous HMM is formed where the state transition probability matrix varies over time (denoted as $P^v(\tau_v)$ here). $\alpha$ and $\beta$ are the forward and backward variables used in the forward-backward algorithm \cite{RabinerIEEE1989}.  }
        \label{fig:inhomog-hmm}
\end{figure}

The key is to note that the posterior state probabilities are only needed at the times where we have observation data. We can exploit this insight to reformulate the estimation problem in terms of an equivalent discrete \emph{time-inhomogeneous} hidden Markov model. This is illustrated in Fig.~\ref{fig:inhomog-hmm}.

Specifically, given the current estimate $\hat Q_0$, $O$ and $T$, we divide the timeline into $V$ intervals, each with duration $\tau_v = t_v - t_{v-1}$. We then make use of the transition property of CTMC, and associate each interval $v$ with a state transition matrix $P^v(\tau_v):=e^{\hat Q_0 \tau_v}$. Together with the emission model $p(o|s)$, this results in a discrete time-inhomogeneous hidden Markov model with joint likelihood:
\begin{align}
  \prod_{v=1}^V [P^v(\tau_v)]_{(s(t_{v-1}), s(t_v))} \prod_{v=0}^V p(o_v | s(t_v)).\label{eq:inHMM}
\end{align}
The formulation in Eq.~\ref{eq:inHMM} allows us to reduce the computation of $p(s(t_v)=k, s(t_{v+1})=l | O, T, \hat{Q}_0)$ to familiar operations. The forward-backward algorithm~\cite{RabinerIEEE1989} can be used to compute the posterior distribution of the hidden states, which we refer to as the \emph{soft} method.  This gives the probabilities $p(s(t_v)=k, s(t_{v+1})=l | O, T, \hat{Q}_0)$, which sum to $1$ over $k$ and $l$.  Alternatively, the MAP assignment of hidden states obtained from the Viterbi algorithm can provide an approximate distribution, which we refer to as the \emph{hard} method.  This gives $p(s(t_v)=k, s(t_{v+1})=l | O, T, \hat{Q}_0)=1$ for a single value of $k$ and $l$, and $p(s(t_v)=k, s(t_{v+1})=l | O, T, \hat{Q}_0)=0$ for all the others. The forward-backward and Viterbi algorithms are then the same as in \cite{RabinerIEEE1989}, except that we replace the transition matrix with $P^v(\tau_{v})$ for each observation.

The \emph{hard} method is potentially faster, but is less accurate in the case of multimodal posteriors.  The \emph{soft} method is more accurate, but requires expectation calculations for every $k$ and $l$.  Note that the \emph{hard} method is only faster when the computation of the end-state conditioned expectations for a single start and end state is less expensive than computing them for all states, which we will see is not always the case.

\subsection{EM Algorithms for CT-HMM}
\label{sec:em}

Pseudocode for the EM algorithm for CT-HMM parameter learning is shown in Algorithm~\ref{alg:em-outerloop}. Multiple variants of the basic algorithm are possible, depending upon the choice of method for computing the end-state conditioned expectations, along with the choice of \emph{hard} or \emph{soft} decoding for obtaining the posterior state probabilities in Eq.~\ref{eq:inHMM}. 

The remaining step in finalizing the EM algorithm is to discuss the computation of the end-state conditioned expectations (ESCE) for $n_{ij}$ and $\tau_i$ from Eqs.~\ref{eq:endn} and~\ref{eq:endt}, respectively.  The first step is to express the expectations in integral form, following~\cite{Hobolth2005}:
\begin{align}
& \mathbb{E}[n_{ij} | s(0)= k, s(t) = l,Q]
=  \frac{q_{i,j}  }{P_{k,l}(t)}
\int_{0}^{t} P_{k,i}(x) P_{j,l}(t-x)\, dx\label{eq:nij}
\\
& \mathbb{E}[\tau_{i} | s(0)= k, s(t) = l, Q]
=  \frac{1}{P_{k,l}(t)}  \int_{0}^{t} P_{k,i}(x) P_{i,l}(t-x)\, dx.\label{eq:taui}
\end{align}
From Eq.~\ref{eq:nij}, we define
$\tau_{k,l}^{i,j}(t) = \int_{0}^{t} P_{k, i}(x) P_{j,l}(t - x)dx = \int_{0}^{t} (e^{Qx})_{k,i} (e^{Q(t-x)})_{j,l}\, dx$, 
while $\tau_{k,l}^{i,i}(t)$ can be similarly defined for Eq.~\ref{eq:taui} (see~\cite{nodelman05} for a related construction). Three primary methods for computing $\tau_{k,l}^{i,j}(t)$ and $\tau_{k,l}^{i,i}(t)$ have been proposed in the CTMC literature: an eigendecomposition based method, which we refer to as \emph{Eigen}, a method called \emph{uniformization} (\emph{Unif}), and a method from~\cite{VanLoan1978} for computing integrals of matrix exponentials, which we call \emph{Expm}.  \emph{Eigen} and \emph{Unif} both involve expressing the terms $P_{k, i}(x) P_{j,l}(t - x)$ as summations and then integrating the summations.  Eigen utilizes an eigendecomposition-based approach, while Unif is based on series approximations.  \emph{Expm} notes a connection between the integrals and a system of differential equations, and solves the system.  We describe each method,  show how to improve the complexity of the \emph{soft Eigen} method, and discuss their tradeoffs.  

\begin{algorithm}[t]
	\caption{CT-HMM Parameter Learning (Soft/Hard)}
	\label{alg:em-outerloop}
	\begin{algorithmic}[1]
		\STATE {\bfseries Input:} data $O=(o_0, ..., o_V)$ and $T=(t_0,\ldots, t_V)$, state set $S$, edge set $L$, initial guess of $Q$ 
		\STATE {\bfseries Output:} transition rate matrix $Q = (q_{ij})$    
		\STATE Find all time intervals between events $\tau_{v}=t_{v+1}-t_{v}$ for $v=1,...,V-1$, $t_{1}=t_{0}=0$
		\STATE Compute  $P(\tau_{v})=e^{Q\tau_{v}}$ for each $\tau_{v}$   
		\REPEAT      
		\STATE Compute $p(v,k,l)=p(s(t_v)=k, s(t_{v+1})=l | O, T, Q)$ for all $v$,  and the complete/state-optimized data likelihood $l$ by using Forward-Backward (soft) or Viterbi (hard) 
		\STATE Use \emph{Expm, Unif} or \emph{Eigen} method to compute $\mathbb{E}[n_{ij}|O, T, Q]$ and $\mathbb{E}[\tau_i|O, T, Q]$
		\STATE Update $q_{ij} = \frac{\mathbb{E}[n_{ij}|O, T, Q]}{ \mathbb{E}[\tau_i|O, T, Q]}$, and $q_{ii} = -\sum_{i \neq j} q_{ij}$   
		\UNTIL{likelihood $l$ converges}
	\end{algorithmic}
\end{algorithm}

Across the three methods, the bottleneck is generally matrix operations, particularly matrix multiplication.  Our finding is that with our improvements, \emph{soft Eigen} is the preferred method except in the case of an unstable eigendecomposition.  It is efficient due to having few matrix multiplications and it is accurate due to being a soft method.  We find in our experiments that it is very fast (see Fig. \ref{fig:bar-chart-all}) and that the stability of \emph{Eigen} is usually not a problem when using a uniform initialization with random noise.  However, in the case where \emph{Eigen} is unstable in any iteration, the alternatives are \emph{soft} Expm, which has the advantage of accuracy, and \emph{hard} Unif, which is often faster.  Note that one can switch back to \emph{Eigen} again once the likelihood is increasing.

\subsubsection{The Eigen Method}

The calculation of the ESCE $\tau_{k,l}^{i,i}(t)$ and
$\tau_{k,l}^{i,j}(t)$
can be done in closed-form if $Q$ can be diagonalized via its eigendecomposition (the \emph{Eigen} method of \cite{Metzner2007} and \cite{MetznerJournal2007}). Consider the eigendecomposition
$Q = U D U^{-1}$, 
where the matrix $U$ consists of all eigenvectors associated with the corresponding eigenvalues of $Q$ in the diagonal matrix $D=diag(\lambda_1, ...,\lambda_n)$.  Then we have $e^{Qt} = U e^{Dt} U^{-1}$
and the integral can be written as:
\begin{align}
\tau_{k,l}^{i,j}(t) 
= \sum_{p=1}^{n} U_{kp} U_{pi}^{-1} \sum_{q=1}^{n} U_{jq} U_{ql}^{-1} \Psi_{pq}(t)
\end{align}
where the symmetric matrix $\Psi(t)=[\Psi_{pq}(t)]_{p,q \in S}$ is defined as:
\begin{align}
\Psi_{pq}(t) = \begin{cases}
t e^{t \lambda_p} \text{~~if~} \lambda_p = \lambda_q\\
\frac{e^{t \lambda_p} - e^{t \lambda_q}}{\lambda_p - \lambda_q} \text{~~if~} \lambda_p \neq \lambda_q\label{eq:eigen-method-inner-element}
\end{cases}
\end{align}
We now describe a method for vectorizing the \emph{Eigen} computation, which results in improved complexity in the \emph{soft} case.  Let $V=U^{-1}$, $\circ$ be the Hadamard (elementwise) product, and $V_{i}^{T}$ refer to the $i$th column of $V$, and $U_{j}$ the $j$th row of $U$, then
\begin{align}
\tau_{k,l}^{i,j}(t)&=[U[V^{T}_{i}U_{j}\circ \Psi]V]_{kl}\label{eq:vector-eigen}
\end{align}
This allows us to perform only one matrix construction for all $k,l$, but still requires two matrix multiplications for each $ij$ with an allowed transition or edge.\footnote{Note that a version of equation \ref{eq:vector-eigen} appears in~\cite{Metzner2007}, but that version contains a small typographic error.}  Appendix A gives a derivation of this result, which was not in the original paper.

We now show how to reuse the matrix-matrix products across edges and replace them by a Hadamard product to improve efficiency further.  A similar idea was explored in~\cite{mcgibbon2015efficient}, but their derivation is for the gradient calculation of a CTMC, which we extend to EM for CT-HMMs.  The intuition is that since matrix multiplication is expensive, by rearranging matrix operations, we can do one matrix multiplication, cache it, and reuse it so that we only do elementwise matrix products for every possible transition combination $i$ and $j$, rather than doing matrix multiplications for every such combination.

Let $F$ be a matrix given by $F_{kl}=\frac{p(s(t_{v})=k,s(t_{v+1})=l|O,T,\hat{Q}_{0})}{P_{kl}(t)}$, and let $A^{ij}=V^{T}_{i}U_{j}\circ \Psi$.  Then
\begin{align}
\sum_{k,l=1}^{|S|}p(s(t_{v})=k,s(t_{v+1})=l|&O,T,\hat{Q}_{0})\mathbb{E}[n_{ij}|s(t_{v})=k,s(t_{v+1})=l,\hat{Q}_{0}]\\
&=q_{ij}\sum_{k,l=1}^{|S|}([U[V^{T}_{i}U_{j}\circ \Psi]V]\circ F)_{kl}\\
&=q_{ij}\sum_{k,l=1}^{|S|}([UA^{ij}V]\circ F)_{kl}
\end{align}
Now note these two properties of the trace and Hadamard product, which hold for any matrices $A,B,C,D$:
\begin{align}
\sum_{ij}(A\circ B)_{ij}&=\text{Tr}(AB^{T})\\
\text{Tr}(ABCD)&=\text{Tr}(BCDA)
\end{align}
Then
\begin{align}
\sum_{k,l=1}^{|S|}([UA^{ij}V]\circ F)_{kl}&=\text{Tr}(UA^{ij}VF^{T})\\
&=\text{Tr}(A^{ij}VF^{T}U)\\
&=\sum_{kl}(A^{ij}\circ (VF^{T}U)^{T})_{kl}\\
&=\sum_{kl}(A^{ij}\circ \underbrace{(U^{T}FV^{T})}_{\text{reuse}})_{kl}\label{eq:reuse}
\end{align}
The term $U^{T}FV^{T}$ is not dependent on $i,j$: only $A^{ij}$ is, and $A^{ij}$ does not require any matrix products to construct.  Thus for each event or time interval, the na\"ive implementation of Eq. \ref{eq:vector-eigen} required two matrix products for each $i,j$ that form an edge. Through the preceding construction, we can reduce this to only two matrix products in total.  We replaced all the other matrix products with Hadamard products.  This improves the complexity by a factor of $S$, the number of states.  The case for the ESCE of the duration $\tau_{i}$ is similar.  Letting $A^{i}=V_{i}^{T}U_{i}\circ\Psi$ and using the subscript $v$ to denote the matrix constructed for observation $v$, the final expectations are
\begin{align}
E[n_{ij}|O,T,\hat{Q}_{0}]&=q_{ij}\sum_{v=1}^{V-1}\sum_{kl}(A^{ij}_{v}\circ (U^{T}F_{v}V^{T}))_{kl}\label{eq:efficientN}\\
E[\tau_{i}|O,T,\hat{Q}_{0}]&=\sum_{v=1}^{V-1}\sum_{kl}(A^{i}_{v}\circ (U^{T}F_{v}V^{T}))_{kl}\label{eq:efficientT}
\end{align}
Note that the \emph{Hard} eigen method avoids explicitly summing over all $k$ and $l$ states. The key matrix manipulation then is the construction of matrices where the rows and columns correspond to $k$ and $l$, respectively. Therefore, \emph{hard} eigen has the same complexity as \emph{soft} eigen when it is formulated as in Eqs.~\ref{eq:efficientN} and~\ref{eq:efficientT}. Thus, \emph{soft} eigen is the preferred choice.

{\bf Computing the ESCE using the \emph{Eigen} method} Algorithm \ref{alg:scale-eigen} presents pseudocode for our \emph{Eigen} method using the Hadamard product and trace manipulations.  The algorithm does two matrix multiplications for each observation, and does only Hadamard products for each state and edge after that.

\begin{algorithm}
\caption{Eigen Algorithm for ESCE}
\label{alg:scale-eigen}
\begin{algorithmic}[1]
\STATE Perform eigendecomposition $Q=UDV$ using balancing, where $V=U^{-1}$
\FOR{$v=1$ to $V-1$}
\STATE Compute $\tau_{v}=t_{v+1}-t_{v}$, set $t=\tau_{v}$
\STATE Compute $\Psi$ with $t=\tau_{v}\Rightarrow O(S^{2})$
\STATE Compute $F_{k,l}=\frac{p(s(t_{v})=k,s(t_{v+1})=l|O,T,\hat{Q}_{0})}{P(t)}$
\STATE Compute $B=U^{T}FV^{T}$
\FOR{each state $i$ in S}
\STATE $A=V_{i}^{T}U_{i}\circ \Psi$
\STATE $E[\tau_{i}|O,T,Q]+=\sum_{k,l=1}^{|S|}(A\circ B)_{kl}\Rightarrow O(S^{2})$
\ENDFOR
\FOR{each edge $(i,j)$ in $L$}
\STATE $A=V_{i}^{T}U_{j}\circ \Psi\Rightarrow O(S^{2})$
\STATE $\mathbb{E}[n_{ij}|O,T,Q]+=q_{ij}\sum_{k,l=1}^{|S|}(A\circ B)_{kl}\Rightarrow O(S^{2})$
\ENDFOR 
\ENDFOR
\end{algorithmic}
\end{algorithm}

\subsubsection*{Stability of the Eigen Method}
In general, \emph{soft Eigen} is the fastest soft method, but $Qt$ can suffer from an ill-conditioned eigendecomposition which can prevent the method from being usable.
In prior CTMC works~(\cite{Metzner2007,MetznerJournal2007}), they mention that the eigendecomposition can potentially be ill-conditioned, but do not characterize the scope of this problem, which we discuss now in more detail. Both the eigenvalue and eigenvector estimation problems can be ill-conditioned. For the eigenvalue problem, the primary issue is the condition number of the eigenvector matrix. This follows from the Bauer-Fike Theorem~(\cite{bauer1960norms}), which gives a bound on the error in estimating the eigenvalues (as a result of a perturbation $\Delta Q$ of the $Q$ matrix):
\begin{align}
\min_{\lambda\in \lambda(Q)}|\lambda-\mu|&\leq ||U||\cdot ||\Delta Q|| \cdot||U^{-1}||\\
&=\kappa(U)||\Delta Q||.
\end{align}
The error between an eigenvalue $\mu$ of $Q+\Delta Q$ and the true eigenvalue $\lambda$ is bounded by the matrix norm of the perturbation, $||\Delta Q||$, and the condition number $\kappa(U)$ of the eigenvector matrix $U$ of $Q$. We now discuss the impact of each of these two terms. The perturbation of $Q$, $||\Delta Q||$, is often due to rounding error and thus depends on the norm of $Q$.  A class of methods known as balancing or diagonal scaling due to~\cite{osborne1960pre} can help reduce the norm of $Q$. In our experiments, balancing did not provide a significant improvement in the stability of the eigendecomposition, leading us to conclude that rounding error was not a major factor. The condition number $\kappa(U)$ captures the structural properties of the eigenvector matrix. We found empirically that certain pathological structures for the $Q$ matrix, such as sparse triangular forms, can produce poor condition numbers. We recommend initializing the $Q$ matrix at the start of EM with randomly-chosen values in order to prevent the inadvertent choice of a poorly-conditioned $U$. We found that uniform initialization, in particular, was problematic, unless random perturbations were added.

Having discussed the eigenvalue case, we now consider the case of the eigenvectors.  For an individual eigenvector $r_{j}$, the estimation error takes the form
\begin{align}
\Delta{r}_{j}&=\sum_{k\neq j}\frac{l_{k}\Delta{Q}r_{j}}{\lambda_{j}-\lambda_{k}}r_{k}+O(||\Delta Q||^{2}),
\end{align}
where $l_{k}$ are left eigenvectors, $r_{j}$ and $r_{k}$ are right eigenvectors, and $\lambda_{j},\lambda_{k}$ are eigenvalues of $Q$ (see \cite{bindel2009principles} for details). Thus the stability of the eigenvector estimate degrades when eigenvalues are closely-spaced, due to the term $\lambda_{j}-\lambda_{k}$ in the denominator. Note that this condition is problematic for the ESCE computation as well, as can be seen in Eq.~\ref{eq:eigen-method-inner-element}. As was the case for the eigenvalue problem, care should be taken in initializing $Q$. 

In summary, we found that randomly initializing the $Q$ matrix was sufficient to avoid problems at the start of EM. While it is difficult in general to diagnose or eliminate the possibility of stability problems during EM iterations, we did not encounter any significant problems in using the \emph{Eigen} approach with either a uniform perturbation plus random noise or simply random initializations in our experiments. We recommend monitoring for a decrease in the likelihood and switching to an alternate method for that iteration in the event of a problem.  One can switch back to \emph{Eigen} once the likelihood is increasing again.

\subsubsection{Expm Method}

Having described an eigendecomposition-based method for computing the ESCE, we now describe an alternative approach based on a classic method of~(\cite{VanLoan1978}) for computing integrals of matrix exponentials. In this approach, an auxiliary matrix $A$ is constructed as 
$A = \begin{bmatrix} Q & B\\ 0 & Q \\ \end{bmatrix}$, where $B$ is a matrix with identical dimensions to $Q$. It is shown in~\cite{VanLoan1978} that
\begin{align}
\int_0^t e^{Qx} B e^{Q(t-x)} dt = (e^{At})_{(1:n), (n+1):(2n)}
\end{align}
where $n$ is the dimension of $Q$.  That is, the integral evaluates to the upper right quadrant of $e^{At}$.  Following~\cite{Hobolth2011}, we set $B = I(i, j)$, where $I(i,j)$ is the matrix with a $1$ in the $(i,j)$th entry and $0$ elsewhere. Thus the left hand side reduces to $\tau_{k,l}^{i,j}(t)$ for all $k, l$ in the corresponding matrix entries, and we can leverage the substantial literature on numerical computation of the matrix exponential. We refer to this approach as \emph{Expm}, after the popular Matlab function.  This method can be seen as expressing the integral as a solution to a differential equation.  See Sec. 4 of  \cite{Hobolth2011} for details.

The most popular method for calculating matrix exponentials is the Pad\'e approximation. As was the case in the \emph{Eigen} method, the two issues governing the accuracy of the Pad\'e approximation are the norm of $Q$ and the eigenvalue spacing. If the norms are large, scaling and squaring, which involves exploiting the identity $e^A=(e^{A/m})^m$ and using powers of two for $m$, can be used to reduce the norm. To understand the role of Eigenvalue spacing, consider that the Pad\'e approximation involves two series expansions $N_{pq}(Qt)$ and $D_{pq}(Qt)$, which are used to construct the matrix exponential as follows:
\begin{align}
e^{Qt}\approx[D_{pq}(Qt)]^{-1}N_{pq}(Qt)
\end{align}
When the eigenvalue spacing \emph{increases}, $D_{pq}(Qt)$ becomes closer to singular, causing large errors as described in~\cite{moler1978nineteen,moler2003nineteen}.

The maximum separation between the eigenvalues is bounded by the Gershgorin Circle Theorem (\cite{golub2012matrix}), which states that all of the eigenvalues of a rate matrix lie in a circle in the complex plane centered at the largest rate, with radius equal to that rate.  That is, all eigenvalues $\lambda\in \lambda(Q)$ lie in $\{z\in \mathbb{C}:|z-\max{q_{i}}|\leq \max{q_{i}}\}$.  
This construction allows us to bound the maximum eigenvalue spacing of $Qt$ (considered as a rate matrix). Two eigenvalues cannot be further apart than twice the absolute value of the largest magnitude diagonal element.  Further, scaling and squaring helps with this issue, as it reduces the magnitude of the largest eigenvalue.  Additional details can be found in~(\cite{moler1978nineteen,moler2003nineteen,Higham2008}).

Because scaling and squaring can address stability issues associated with the Pad\'e method, we conclude that \emph{Expm} is the most stable method for computing the ESCE.  However, we find it to be dramatically slower than \emph{Eigen} (especially given our vectorization and caching improvements), and so it should only be used if \emph{Eigen} fails.

The \emph{Expm} algorithm does not have an obvious hard variant.  Hard variants involve calculating expectations conditioned on a single start state $k$ and end-state $l$ for the interval between the observations.  However, \emph{Expm}, by virtue of using the Pad\'e approximation of the matrix exponential, calculates it for all $k$ and $l$.  The output of the matrix exponential gives a matrix where each row corresponds to a different $k$ and each column a different $l$.  Developing a hard variant would thus require a method for returning a single element of the matrix exponential more efficiently than the entire matrix.  One direction to explore would be the use of methods to compute the action of a matrix exponential $e^{At}x$, where $x$ is a vector with a single $1$ and $0$'s elsewhere, without explicitly forming $e^{At}$ (see~\cite{al2011computing}).

{\bf Computing the ESCE using the \emph{Expm} method}
Algorithm \ref{alg:soft-soft-expm} presents pseudocode for the \emph{Expm} method for computing end-state conditioned expectations. The algorithm exploits the fact that the $A$ matrix does not change with time $t$. Therefore, when using the \emph{scaling and squaring} method~(\cite{Higham2008}) for computing matrix exponentials, one can easily cache and reuse the intermediate powers of $A$ to efficiently compute $e^{At}$ for different values of $t$.

\begin{algorithm}[ht]
	\small
	\caption{Expm Algorithm for ESCE}
	\label{alg:soft-soft-expm}
	\begin{algorithmic}[1]
    	\FOR{$v=1$ {\bfseries to} $V-1$}
        \STATE $\tau_{v}=t_{v+1}-t_{v}$, set $t=\tau_{v}$
		\FOR{each state $i$ in $S$}
		\STATE $D_i = \frac{(e^{At})_{(1:n), (n+1):(2n)}}{P_{kl}(t)}$, where \small{$A = \begin{bmatrix} Q & I(i,i) \\ 0 & Q \\ \end{bmatrix}$ }

		\STATE $\mathbb{E}[\tau_i|O, T, Q]~+=~ \sum_{(k,l) \in L} p(s(t_{v})=k,s(t_{v+1})=l|O,T,\hat{Q}_{0})  (D_i)_{k,l} $ 
		\ENDFOR
		\FOR{each edge ($i,j$) in $L$}
		\STATE $N_{ij} = \frac{q_{ij} (e^{At})_{(1:n), (n+1):(2n)}}{P_{kl}(t)}$, where \small{$A = \begin{bmatrix} Q & I(i,j) \\ 0 & Q \\ \end{bmatrix}$ } 

		\STATE $\mathbb{E}[n_{ij}|O, T, Q]~ += \sum_{(k,l) \in L} p(s(t_{v})=k,s(t_{v+1})=l|O,T,\hat{Q}_{0})  (N_{ij})_{k,l} $ 
		\ENDFOR
		\ENDFOR  
	\end{algorithmic}
\end{algorithm}

\subsubsection{Uniformization}
We now discuss a third approach for computing the ESCE.  This was first introduced by ~\cite{Hobolth2011} for the CTMC case, and is called \emph{uniformization} (\emph{Unif}).  \emph{Unif} is an efficient approximation method for computing the matrix exponential $P(t) = e^{Qt}$(~\cite{Jensen1953,Hobolth2011}). It gives an alternative description of the CTMC process and illustrates the relationship between CTMCs and DTMCs (see \cite{RossBook1983}).  The idea is that instead of describing a CTMC by its rate matrix, we can subdivide it into two parts: a Poisson process $\{N(t):t\geq 0\}$ with mean $\hat{q}$, where $N(t)$ refers to the number of events under the Poisson process at time $t$, and a DTMC and its associated transition matrix $R$.  The state of the CTMC at time $t$ is then equal to the state after $N(t)$ transitions under the DTMC transition matrix $R$.  In order to represent a CTMC this way, the mean of the Poisson process and the DTMC transition matrix must be selected appropriately.

Define $\hat{q} = \max_i q_i$, and matrix $R = \frac{Q}{\hat{q}} + I$, where $I$ is the identity matrix. Then,
\begin{align}e^{Qt} = e^{\hat{q}(R-I)t} = \sum_{m=0}^{\infty} R^m \frac{(\hat{q} t)^m}{m!} e^{-\hat{q} t} 
= \sum_{m=0}^{\infty} R^m Pois(m ; \hat{q} t),
\end{align}
where $Pois(m; \hat{q} t)$ is the probability of $m$ occurrences from a Poisson distribution with mean $\hat{q} t$.  The expectations can then be obtained by directly inserting the $e^{Qt}$ series into the integral: 
\begin{align}
\tau^{i,i}_{k,l} &= \frac{\sum_{m=0}^{\infty} \frac{t}{m+1} [ \sum_{n=0}^{m} (R^n)_{ki} (R^{m-n})_{il} ] Pois(m; \hat{q} t)}{P_{kl}(t)}\\
\tau^{i,j}_{k,l} &= \frac{R_{ij}
\sum_{m=1}^{\infty}   [ \sum_{n=1}^{m} (R^{n-1})_{ki}  (R^{m-n})_{jl} ] Pois(m; \hat{q} t)}{P_{kl}(t)}
\end{align}
The main difficulty in using \emph{Unif} in practice lies in determining the truncation point for the infinite sum. However, for large values of $\hat{q} t$, we have $Pois(\hat{q} t) \approx \mathcal{N}(\hat{q} t, \hat{q} t)$, where $\mathcal{N}(\mu, \sigma^2)$ is the normal distribution and one can then bound the truncation error from the tail of Poisson by using the cumulative normal distribution~(\cite{TataruBio2011}). Our implementation uses a truncation point at $M = \ulcorner 4 + 6 \sqrt{\hat{q} t} + (\hat{q} t) \urcorner$, which is suggested in~(\cite{TataruBio2011}) to have error bound of $10^{-8}$.

{\bf Computing the ESCE using the \emph{Unif} method} Algorithm~\ref{alg:unif} presents pseudocode for the \emph{Unif} method for computing end-state conditioned expectations. The main benefit of \emph{Unif} is that the $R$ sequence ($R,  R^2,...,R^{\hat{M}}$) can be precomputed (line 2) and reused, so that no additional matrix multiplications are needed to obtain all of the expectations. One main property of \emph{Unif} is that it can evaluate the expectations for only the two specified end-states, and it has $O(M^2)$ complexity, which is not related to $S$ (when given the precomputed $R$ matrix series).

One downside of \emph{Unif} is that if $\hat{q_i} t$ is very large, so is the truncation point $M$.  The computation can then be very time consuming. We find that \emph{Unif}'s running time performance depends on the data and the underlying Q values. The time complexity analysis is detailed in Algorithm \ref{alg:unif} line 15.  This shows that the complexity of \emph{soft Unif} is unattractive, while \emph{hard Unif} may be attractive if \emph{Eigen} fails due to instability.

\begin{algorithm}[!ht]
	\caption{Unif Algorithm for ESCE}  
	\label{alg:unif}
	\begin{algorithmic}[1]
		\STATE Set $\hat{t} = max ~t_\Delta$; set $\hat{q} = max_i q_i$.
		\STATE Let $R = Q/\hat{q} + I$. Compute $R, R^2,...,R^{\hat{M}}$, $\hat{M} = \ulcorner 4 + 6 \sqrt{\hat{q} \hat{t}} + (\hat{q} \hat{t}) \urcorner$ $\Rightarrow O(\hat{M} S^3)$

		\FOR{$v=1$ {\bfseries to} $V-1$}
        \STATE $\tau_{v}=t_{v+1}-t_{v}$, set $t = \tau_{v}$
		\STATE $M = \ulcorner 4 + 6 \sqrt{\hat{q} t } + (\hat{q} t) \urcorner$; 
		\FOR{each state $i$ in $S$}
		\STATE $ E[\tau_i | s(0) = k, s(t) = l, Q] =
		\frac{\sum_{m=0}^M \frac{t}{m+1} [ \sum_{n=0}^{m} (R^n)_{ki} (R^{m-n})_{il} ] Pois(m; \hat{q} t)}{P_{kl}(t)}  $
		$\Rightarrow O(M^2)$
		\STATE  $E[\tau_i|O, T, Q] +=p(s(t_{v})=k,s(t_{v+1})=l|O,T,\hat{Q}_{0}) E[\tau_i | s(0) = k, s(t) = l] $   
		\ENDFOR           
		\FOR{each edge $(i,j)$ in $L$}      
		\STATE $ E[n_{ij} | s(0) = k, s(t) = l, Q] = \frac{R_{ij} \sum_{m=1}^M   [ \sum_{n=1}^{m} (R^{n-1})_{ki}  (R^{m-n})_{jl} ] Pois(m; \hat{q} t) }{P_{kl}(t)}
		$  $\Rightarrow O(M^2)$   
		\STATE $E[n_{ij} | O, T, Q] += p(s(t_{v})=k,s(t_{v+1})=l|O,T,\hat{Q}_{0}) E[n_{ij} | s(0) = k, s(t) = l]$ 
		\ENDFOR
		\ENDFOR
		\STATE 		
		{\footnotesize \emph{Soft}: $O(\hat{M} S^3 + V S^3 M^2 + VS^2 L M^2)$;
		\emph{Hard}: $O(\hat{M} S^3 + V S M^2 + V L M^2)$ }
	\end{algorithmic} 
\end{algorithm}

\subsubsection{Summary of Time Complexity}

To compare the computational cost of different methods, we conducted an asymptotic complexity analysis for the five combinations of \emph{hard} and \emph{soft} EM with the methods \emph{Expm}, \emph{Unif}, and \emph{Eigen} for computing the ESCE. The complexities are summarized in Table~\ref{table:time_complexity}. \emph{Eigen} is the most attractive of the soft methods at $O(VS^3 + VLS^2)$, where $V$ is the number of visits, $S$ is the number of  states, and $L$ is the number of edges.  Its one drawback is that the eigendecomposition may become ill-conditioned at any iteration.  However, in our experiments, with a random initialization, we found Eigen to be successful, and other papers have found similar results \cite{Metzner2007}, although generally with a smaller number of states.  If \emph{Eigen} fails, \emph{Expm} provides an alternative soft method, and \emph{Unif} provides an alternative hard method. \emph{Hard Unif} is often faster than \emph{Expm} in practice, so we recommend running that first to get a sense of how long \emph{Expm} will take, and if it is feasible, run \emph{Expm} afterwards.

\begin{table*}[t]
\small
	\centering	
	\newcommand{\tabincell}[2]{\begin{tabular}{@{}#1@{}}#2\end{tabular}}
	\caption{\small
		Time complexity comparison of all methods in evaluating all required expectations under \emph{Soft/Hard} EM,
		$S$: number of states, $L$: number of edges, $V$: number of visits, $M$: the largest truncation point of the infinite sum for \emph{Unif}, set as $\ulcorner 4 + 6 \sqrt{\hat{q} \hat{t} } + (\hat{q} \hat{t}) \urcorner$, where $\hat{q}= \max_i q_i$, and $\hat{t} = max_{\Delta} \tau_{v}$).
	}
	\begin{tabular}{|l||l|l|l|}
		\hline
		complexity & Expm & Unif & Eigen\\ \hline \hline
		Soft EM & $O(V S^4 + V L S^3)$  & \tabincell{c}{$O(M V^3 + V S^3 M^2$\\$ + V S^2 L M^2)$} & $O(VS^{3}+VLS^{2})$\\ \hline
		Hard EM & $O(V S^4 + V L S^3)$ &		
		\tabincell{c}{     
        $O(M S^3 + V S M^2$ \\  $+V L M^2)$} & N/A\\ \hline		
	\end{tabular}	 
	\label{table:time_complexity}
\end{table*}

The time complexity comparison between \emph{Expm} and \emph{Unif} depends on the relative size of the state space $S$ and $M$, where $M = \ulcorner 4 + 6 \sqrt{\max_i q_i t } + (\max_i q_i t) \urcorner$ is the largest truncation point of the infinite sum used in \emph{Unif} (see Table~\ref{table:time_complexity}). It follows that \emph{Unif} is more sensitive to $\max_i q_i t$ than \emph{Expm} (quadratic versus log dependency). This is because when \emph{Expm} is evaluated using the scaling and squaring method \cite{Higham2008}, the number of matrix multiplications depends on the number of applications of matrix scaling and squaring, which is $\ulcorner \log_2(||Q t||_1/\theta_{13})\urcorner$, where $\theta_{13}=5.4$ (the \emph{Pade} approximant  with degree 13).  If scaling of $Q$ is required \cite{Higham2008}, then we have $\log_2(||Q t||_1) \leq \log_2(S \max_{i} q_i t)$. Therefore, the running time of \emph{Unif} will vary with $\max q_i  t$ more dramatically than \emph{Expm}. 


When selecting an EM variant, there are three considerations: stability, time, and accuracy. Overall, \emph{soft Eigen} offers the best tradeoff between speed and accuracy. However, if it is not stable, then \emph{soft Expm} will generally have higher accuracy than \emph{hard Unif}, but may be less efficient.

In some applications, event times are distributed irregularly over a discrete timescale.  For example, hospital visits may be identified by their date but not by their time.  In that case, the interval between two events will be a discrete number.  In such cases, events with the same interval, e.g. with five days between visits, can be pooled and the ESCE can be computed once for all such events.  See \cite{LiuNips15} for details, including the supplementary material for complexity analysis.

\subsection{Experimental results}
\label{sec:exp}

We evaluated our EM algorithms in simulation (Sec.~\ref{sec:exp1}) and on two real-world datasets: a glaucoma dataset (Sec.~\ref{sec:exp2}) in which we compare our prediction performance to a state-of-the-art method, and a dataset for Alzheimer's disease (AD, Sec.~\ref{sec:exp3}) where we compare visualized progression trends to recent findings in the literature. Our disease progression models employ 105 (Glaucoma) and 277 (AD) states, representing a significant advance in the ability to work with large models (previous CT-HMM works~\cite{JacksonJSS2011,MurilloNIPS2011Complete,Wang2014kdd} employed fewer than 100 states).  We initialized the rate matrix uniformly with a small random perturbation added to each element. The random perturbation avoids a degenerate configuration for the \emph{Eigen} method, while uniform initialization makes the runtimes comparable across methods. We used balancing for the eigendecomposition. Our timing experiments were run on an early 2015 MacBook Pro Retina with a 3.1 GHz Intel Core i7 processor and 16 GB of memory.



\subsubsection{Simulation on a 5-state Complete Digraph}
\label{sec:exp1}

We test the accuracy of all methods on a 5-state complete digraph with synthetic data generated under different noise levels.
Each $q_i$ is randomly drawn from $[1,5]$ and then $q_{ij}$ is drawn from $[0,1]$ and renormalized such that $  \sum_{j \neq i} q_{ij} = q_i$. The state chains are generated from $Q$, such that each chain has a total duration around $T = \frac{100}{\min_i q_i} $, where $\frac{1}{\min_i q_i}$ is the largest mean holding time. The data emission model for state $i$ is set as $N(i, \sigma^2)$, where $\sigma$ varies under different noise level settings.
The observations are then sampled from the state chains with rate $\frac{0.5}{\max_i q_i}$, where $\frac{1}{\max_i q_i}$ is the smallest mean holding time, which should be dense enough to make the chain identifiable. A total of $10^5$ observations are sampled. 
The average 2-norm relative error 
$\frac{||\hat{q} - q||}{||q||}$ is used as the performance metric, 
where $\hat{q}$ is a vector contains all learned $q_{ij}$ parameters, and  $q$ is the ground truth.


The simulation results from 5 random runs are given in Table~\ref{table:learn_exp2}. \emph{Expm}, \emph{Unif}, and \emph{Eigen} produced nearly identical results, and so they are combined in the table, which focuses on the difference between hard and soft variants. We found \emph{Eigen} to be stable across all runs. All soft methods achieved significantly higher accuracy than hard methods, especially for higher observation noise levels. This can be attributed to the maintenance of the full hidden state distribution, leading to improved robustness to noise.

\begin{table*}[!t]
\small
\centering		
	\caption{ 
	\small	
		The average 2-norm relative error 
		from 5 random runs on a 5-state complete digraph under varying measurement noise levels. 
	}\vskip -0.1in	
	\begin{tabular}{|l||l|l|l|l|l|}
		\hline
		\footnotesize
		Error & $\sigma=1/4$ & $\sigma=3/8$ & $\sigma=1/2$ & $\sigma=1$ & $\sigma=2$ \\ \hline \hline
		Soft & 0.026$\pm$0.008 &  0.032$\pm$0.008 &  0.042$\pm$0.012 & 0.199$\pm$0.084 & 0.510$\pm$0.104 \\ \hline
		Hard & 0.031$\pm$0.009 &  0.197$\pm$0.062 &  0.476$\pm$0.100 & 0.857$\pm$0.080 & 0.925$\pm$0.030 \\ \hline			
	\end{tabular}
	\label{table:learn_exp2}
\end{table*}

\begin{figure}[tb]
	\vskip -0.1in
	\centering
	\includegraphics[width=14cm]{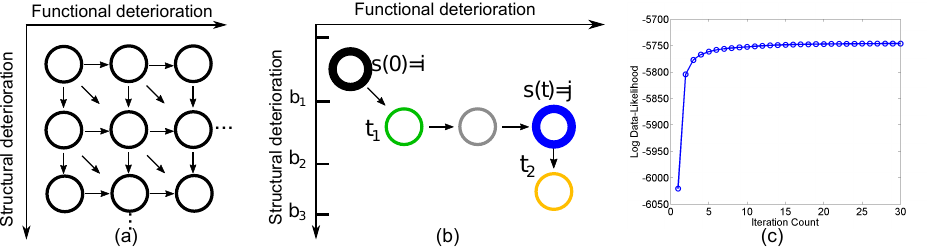}
	\vskip -0.05in
	\caption{	\small
		(a) The 2D-grid state structure for glaucoma progression modeling. (b) Illustration of the prediction of future states from $s(0)=i$.
		(c) One fold of convergence behavior of \textit{Soft(Expm)} on the glaucoma dataset.
	}
	\label{fig:glaucoma}
	\vskip -0.15in
\end{figure}

In the next set of experiments, we used the CT-HMM to analyze and visualize disease progression patterns from two real-world datasets of glaucoma and Alzheimer's Disease (AD), and show predictive performance on the Glaucoma dataset. Both are examples of degenerative disorders where the time course of the disease plays an important role in its etiology and treatment. We demonstrate that CT-HMM can yield insight into disease progression, and we compare the timing results for learning across our family of methods.

\subsubsection{Application of CT-HMM to Predicting Glaucoma Progression}
\label{sec:exp2}

\begin{figure}[ht]
	\centering
	\ifx\includegraphics[height=3cm]{2D_glaucoma_model_struct.pdf}\hfill\fi
	\includegraphics[height=5.7cm]{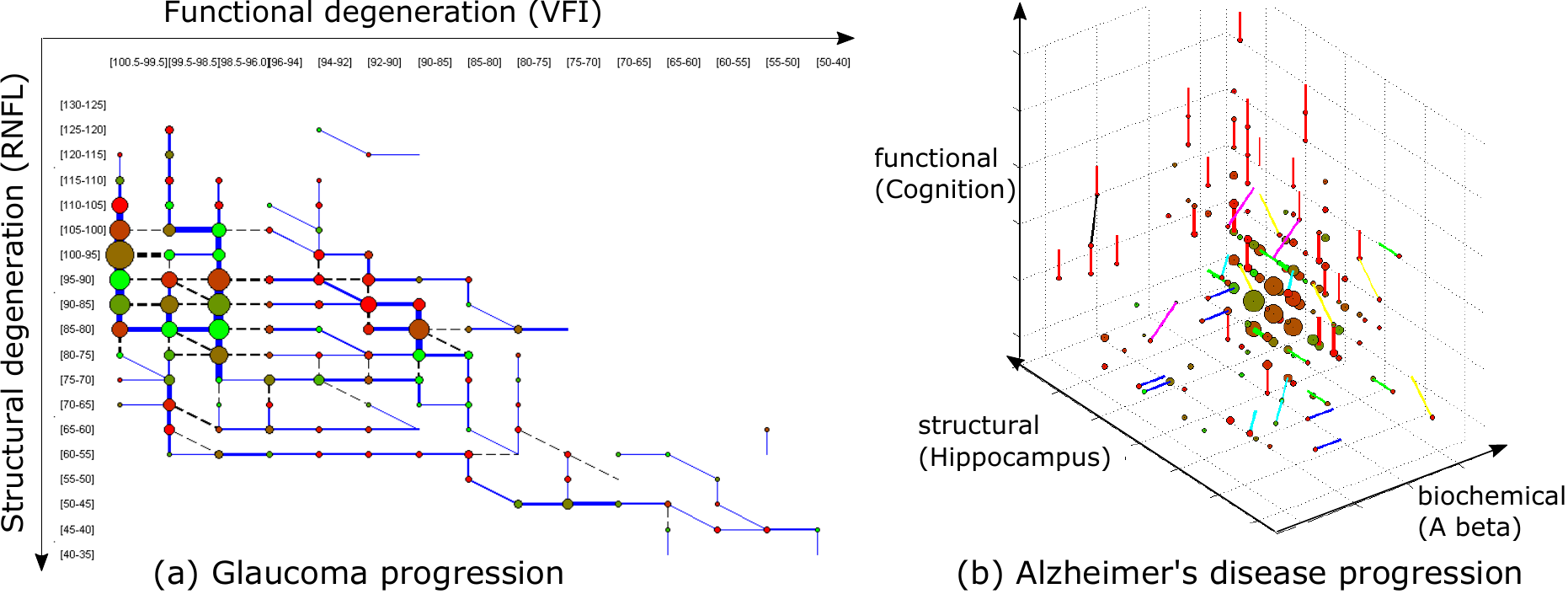}
	\vskip -0.05in
	\caption{Visualization scheme: (a) Nodes represent states of glaucoma, with the node color encoding the average sojourn time (red to green: 0 to 5 years and above). The blue links between nodes indicate the most probable (i.e. strongest) transitions between adjacent states, selected from among the three allowed transitions (i.e., down, to the right, and diagonally). The line width and the node size reflect the expected count of patients passing through a transition or state. (b) The representation for AD is similar to (a) with the strongest transition from each state being coded as follows: $A\beta$ direction
(blue), hippo (green), cog (red), $A\beta$+hippo (cyan), $A\beta$+cog (magenta), hippo+cog (yellow), $A\beta$+hippo+ cog(black). The node color represents the average sojourn time (red to green: 0 to 3 years and above).}
	\label{fig:glaucoma-alzheimers}
\end{figure}

We now describe a 2D CT-HMM for glaucoma progression. Glaucoma is a leading cause of blindness and visual morbidity worldwide (\cite{KingmanWHO2004}). This disease is characterized by a slowly progressing optic neuropathy with associated irreversible structural and functional damage. We use a 2D-grid state space model defined by successive value bands of the two main glaucoma markers, Visual Field Index (VFI) (functional marker) and average RNFL (Retinal Nerve Fiber Layer) thickness (structural marker) with forwarding edges (see Fig. \ref{fig:glaucoma-alzheimers}(a)).

Our glaucoma dataset contains 101 glaucomatous eyes from 74 patients followed for an average of 11.7$\pm$4.5 years, and each eye has at least 5 visits (average 7.1$\pm$3.1 visits). There were 63 distinct time intervals. The state space is created so that most states have at least 5 raw measurements mapped to them. All states that are in a direct path between two successive measurements are instantiated, resulting in 105 states.

To predict future continuous measurements, we follow a simple procedure illustrated in Fig.~\ref{fig:glaucoma}(b). Given a testing patient, Viterbi decoding is used to decode the best hidden state path for the past visits. Then, given a future time $t$, the most probable future state is predicted by $j = \max_j P_{ij}(t)$ (blue node), where $i$ is the current state (black node). To predict the continuous measurements, 
we search for the future time $t_1$ and $t_2$ in a desired resolution when the patient enters and leaves a state having same value range as state $j$ for each disease marker separately.
The measurement at time $t$ can then be computed by linear interpolation between $t_1$ and $t_2$ and the two data bounds of state $j$ for the specified marker ($[b1,b2]$ in Fig. \ref{fig:glaucoma}(b)). 
The mean absolute error (MAE) between the predicted values and the actual measurements was used for performance assessment. The performance of CT-HMM was compared to both conventional linear regression and Bayesian joint linear regression~\cite{Medeiros2012}. 
For the Bayesian method, 
the joint prior distribution of the four parameters (two intercepts and two slopes) computed from the training set \cite{Medeiros2012} is used alongside the data likelihood. The results in Table~\ref{table:predict} demonstrate the significantly improved performance of CT-HMM.

In Fig.~\ref{fig:glaucoma-alzheimers}(a), we visualize the model trained using the entire glaucoma dataset. Several dominant paths can be identified: there is an early stage containing RNFL thinning with intact vision (blue vertical path in the first column). At RNFL range $[80,85]$, the transition trend reverses and VFI changes become more evident (blue horizontal paths). This $L$ shape in the disease progression supports the finding in~\cite{Wollstein2012} that RNFL thickness of around $77$ microns is a tipping point at which functional deterioration becomes clinically observable with structural deterioration. Our 2D CT-HMM model reveals the non-linear relationship between structural and functional degeneration, yielding insights into the progression process.

\vskip -0.12in
\begin{table*}[htb]
\small
	\centering	
	\caption{ \small
		The mean absolute error (MAE) of predicting the two glaucoma measures.
		($*$ represents that CT-HMM performs significantly better than the competing method under \textit{student t-test}).
	}
	\vskip -0.1in
	\begin{tabular}{|c||c|c|c|}
		\hline
		MAE & CT-HMM  & Bayesian Joint Linear Regression & Linear Regression \\ \hline \hline
		VFI & 4.64 $\pm$ 10.06 & 5.57 $\pm$ 11.11 * ($p=0.005$)
		&  7.00 $\pm$ 12.22  *($p \approx 0.000$) \\ \hline
		RNFL & 7.05 $\pm$ 6.57 &  9.65 $\pm$ 8.42 * ($p \approx 0.000$)
		&  18.13 $\pm$ 20.70 * ($p \approx 0.000$) \\ \hline
	\end{tabular}
	\label{table:predict}
\end{table*}


\subsubsection{Application of CT-HMM to Exploratory Analysis of Alzheimer's Disease}

\label{sec:exp3}
We now demonstrate the use of CT-HMM as an exploratory tool to visualize the temporal interaction of disease markers of Alzheimer's Disease (AD). AD is an irreversible neuro-degenerative disease that results in a loss of mental function due to the degeneration of brain tissues. An estimated 5.3 million Americans have AD, yet no prevention or cures have been found~\cite{ADNIweb}. It could be beneficial to visualize the relationship between clinical, imaging, and biochemical markers 
as the pathology evolves, in order to better understand AD progression and develop treatments.

In this experiment, we analyzed the temporal interaction among the three kinds of markers: amyloid beta ($A\beta$) level in cerebral spinal fluid (CSF) (a bio-chemical marker), hippocampus volume (a structural marker), and ADAS cognition score (a functional marker). We obtained the \emph{ADNI} (The \emph{Alzheimer's Disease Neuroimaging Initiative}) dataset from~\cite{ADNIweb}.\footnote{Data were obtained from the ADNI database (\url{adni.loni.usc.edu}). 
	The ADNI was launched in
	2003 as a public-private partnership, led by Principal Investigator Michael W. Weiner,
	MD. The primary goal of ADNI has been to test whether serial magnetic resonance imaging
	(MRI), positron emission tomography (PET), other biological markers, and clinical and
	neuropsychological assessment can be combined to measure the progression of mild
	cognitive impairment (MCI) and early Alzheimer's disease (AD). For up-to-date information,
	see \url{http://www.adni-info.org}
	}
Our sample included patients with mild cognitive impairment (MCI) and AD who had at least two visits with all three markers present, yielding 206 subjects with an average of $2.38 \pm 0.66$ visits traced in $1.56 \pm 0.86$ years. The dataset contained 3 distinct time intervals at one month resolution. A 3D gridded state space consisting of 277 states with forwarding links was defined such that for each marker, there were 14 bands that spanned its value range. The procedure for constructing the state space and the definition of the data emission model is the same as in the glaucoma experiment. Following CT-HMM learning, the resulting visualization of Alzheimer's disease in Fig.~\ref{fig:glaucoma-alzheimers}(b) supports recent findings that a decrease in the A level of CSF (blue lines) is an early
marker that precedes detectable hippocampus atrophy (green lines) in cognition-normal elderly (\cite{Anne2009}). The CT-HMM disease model and its associated visualization can be used as an exploratory tool to gain insights into health dynamics and generate hypotheses for further investigation by biomedical researchers.

\begin{figure}[ht]
	\centering
	\includegraphics[width=10cm]{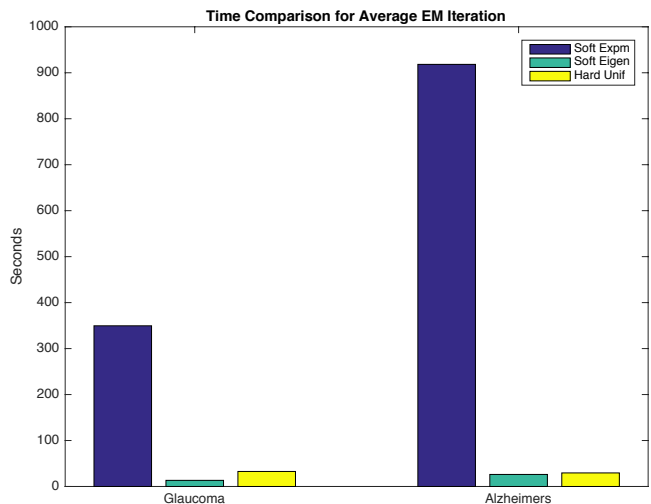}
	\vskip -0.05in
	\caption{Time comparison for the average time per iteration between \emph{soft Expm}, \emph{soft Eigen} and \emph{hard Unif} for both experiments.  \emph{Soft Eigen} is the fastest method, over an order of magnitude faster than \emph{soft Expm} in both cases.  Thus, it should be used unless the eigendecomposition fails, in which case there is a tradeoff between \emph{soft Expm} for accuracy and \emph{hard Unif} for speed.\small
		}
	\label{fig:bar-chart-all}
\end{figure}

Fig.~\ref{fig:bar-chart-all} gives the average runtime comparison for a single EM iteration between \emph{soft Expm}, \emph{soft Eigen}, and \emph{hard Unif} for both datasets.  \emph{Soft Eigen} with our improvements is 26 times faster than \emph{soft Expm} for the glaucoma experiment, and 35 times faster for the AD experiment.  \emph{Hard Unif} is slightly slower than \emph{soft Eigen}.  We did not include \emph{soft Unif} due to its poor complexity or \emph{hard Eigen} due to its minimal computational benefit in comparison to \emph{soft Eigen}.

\section{Trajectory Decoding in Continuous-Time Hidden Markov Models}
\subsection{Introduction to decoding problems in Continuous-time HMM}


\label{sec:intro-decode}
In CT-HMM, decoding the optimal state transition sequence as well as the corresponding dwelling time of each state, given noisy observations, is an important problem for understanding the underlying hidden dynamics. To the best of our knowledge, there is no existing literature solving this problem for CT-HMM. To decode the best state \textit{at} the observation time, there is a known method by forming an equivalent time-inhomegeneous HMM and then using \textit{Viterbi} decoding (see Section \ref{sec:computing-posterior-state}).
However, the problem of finding the optimal state transition trajectory in between the decoded states at observation times for CT-HMM has not been addressed in the literature before. While state transition trajectories can be represented with the CT-HMM's transition rate matrix's transition probabilities and average sojourn times as shown in Fig. 5 and as done by \cite{Kwon2020}, this is not the decoded state transition trajectory as  calculated for each patient based on their individual observations. Here, we present an algorithm that decodes the optimal state transition trajectory from the CT-HMM, which leverages recent literature for end-state conditioned CTMC state sequence decoding and advances in computing the expected state dwelling time. This approach allows us to calculate the individual state trajectories for each patient, allowing for the identification of different subsets of patients based upon their modeled disease progression. We show its application by illustrating the differences between decoded state trajectories in Section 3.4.4 for two different labeled subsets of children with ASD: minimally verbal and verbally developed. 

We will briefly review the prior work for finding the optimal state sequence given two end-states and a total time in CTMC, which is closely related to the decoding for CT-HMM model.
There are two different problem settings for the decoding problems in the CTMC literature, when given two end-states and a total time.
One is to find the most likely state and duration sequence together (\cite{Perkins2009NIPS}), and the other is to find the most probable state sequence but considering all possible state dwelling time assignments for this state sequence (\cite{Levin2012}).
From the analysis in \cite{Perkins2009NIPS} and \cite{Levin2012},
the optimal solution  in the former setting will assign all time to the state with the longest mean holding time (the smallest $q_i$), which is usually undesirable in practice (\cite{Perkins2009NIPS}), while the latter problem setting can result in a more biological meaningful solution (\cite{Levin2012}).




We now discuss the first decoding problem setting for CTMC in more detail.  This involves inferring the most likely state and duration sequence together (also referred to as a trajectory, which is a sequence of states as well as the exact amount of time spent in each state) given two end-states and a total time.
This problem is solved exactly in \cite{Perkins2009NIPS}. 
It is discovered that 
the maximum likelihood trajectory does not exist if a CTMC has a cycle of states $(s_0, s_1, ..., s_k = s_0)$ such that 
$\prod_{i=0}^{k-1} v_{s_i, s_{i+1}} q_{s_i} \geq 1 $ (recall that $v_{ij} = q_{ij}/q_i$).
In this case,
a sequence of trajectories with ever-increasing likelihood can be found if the cycle is reachable, and thus the maximum likelihood trajectory does not exist. The well-defineness of the given Q matrix for finding the maximum likelihood trajectory can be checked using a polynomial time procedure. 
When the well-definedness holds (i.e., no such cycles),  an exact dynamic programming algorithm for inferring the most likely trajectories for CTMCs is designed (\cite{Perkins2009NIPS}). 
However, the optimal solution is found to assign all the time to the state with the longest average holding time (i.e., the state with the smallest $q_i$), and infinitessimal to others. This optimal solution is thus non-representative of typical system behavior (\cite{Perkins2009NIPS}).


The second problem setting we mentioned for CTMC decoding is to find the most probable state sequence while marginalizing out the state dwelling time.  This was recently tackled in \cite{Levin2012}. This setting uses probability rather than likelihood, so there always exists a best solution. \cite{Levin2012} proposed a path extending and pruning algorithm, which is guaranteed to find the optimal solution.
The time complexity of this method depends in a complex way on the underlying state transition matrix $Q$ and the specified total time. Thus, there is currently no easy time complexity analysis of the method, but it is guaranteed to terminate in finite time. This prior work solves the best hidden state transition path given two  end states and a total time. However, it did not present a method to compute the expected state dwelling time for each state in the optimal path given a total time. We believe that both the best state sequence and its corresponding expected state dwelling time are useful for real applications, such as hidden trajectory understanding, and clustering for progression phenotyping.

In the remaining sections, we first review the related work on CTMC decoding  (\cite{Levin2012}) that find the most probable state sequence considering all possible dwelling time assignments.  Then, we present our methods for finding an optimal state transition trajectory in CT-HMM, which leverages the  work from \cite{Levin2012} and recent advances in computing the expected state dwelling time. In the experiments, we show the accuracy of our CT-HMM decoding method under different observation time intervals and noise levels using a synthetic dataset, as well as visualizing the decoded hidden state trajectories of patient data on a real glaucoma dataset. Finally, we will conclude the section apply our proposed decoding methodology to understand the early language development of children with autism spectrum disorder.

\subsection{Related work: search for end-state and total-time conditioned maximum probability state sequences in CTMC}

\label{sec:decode_state_path}

Our decoding algorithm for CT-HMM will need to search for end-state and total-time conditioned maximum probability state sequences in CTMC.
This problem in CTMC is first solved in \cite{Levin2012} which uses a path extending and pruning method.


This problem is formulated as follows (\cite{Levin2012}). Given a CTMC = $(S, Q)$, where $S$ is the state set and $Q$ is the transition matrix, define the time-dependent probability $P_t(G)$ as the probability that the chain visits precisely the sequence of states in $G = (s_1, ..., s_n)$ by time $t$ given that it starts at $s_1$ at time 0, and it is at state $s_n$ at time $t$. Let $\tau_1, \tau_2, ..., \tau_n$ be dwelling time in the corresponding states in $G$ and the total time is $T$. We have:
\begin{align}
P_T(G) = (\prod_{i=1}^n v_{s_i, s_{i+1}}) p(\sum_{i=1}^{n-1} \tau_i \leq T \leq \sum_{i=1}^n \tau_i).
\end{align}
where $v_{s_i, s_j} = q_{ij}/q_i$, is the transition probability between two states.
The goal is to find the most probable $G^*$:
\begin{align}
G^* = max_{~G} ~ P_T(G). 
\end{align}


To evaluate $P_t(G)$, \cite{Levin2012} uses the Chapman-Kolmogorov equations, which results in
a recursive relationship:
$\frac{d P_t(G)}{dt} = q_{s_{n-1}} v_{s_{n-1}, s_n} P_t(\bar{G}) - q_{s_n} P_t(G)$
, where $\bar{G}$ is the state sequence $(s_1, ..., s_{n-1})$, which is the one-step shorter sequence of $G$. 
To evaluate $P_t(G)$, the above linear differential equation needs to be solved. This equation depends on $P_t(\bar{G})$, which obeys its own linear differential equation that also depends on the probability of a one step shorter state sequence.
Thus, $P_t(G)$ can be evaluated by solving a system of linear differential equations, where the variables are $P_t(s_1)$, $P_t((s_1, s_2))$, ..., $P_t((s_1, s_2, ..., s_n))$, with the initial conditions  $P_0(s_1)=1$ and $P_0((s_1,..., s_i))=0, i > 0, i \neq 1$. \cite{Levin2012} uses standard numerical methods to solve these differential equations and calculate $P_t(G)$.

Then, to find the optimal state sequence $G^*$ given a total time $T$, \cite{Levin2012} proposed a way to guide the state path extending and pruning process using a notion called \textit{dominance}. Suppose $G_1$ and $G_2$ are two different state sequences with the same starting and ending state. Then $G_1$ dominates $G_2$ if $P_t(G_1) > P_t(G_2)$ for all $t \in (0, T)$. If $G$ is not dominated by any other sequence, then $G$ is said to be \textit{non-dominated}. 
Two properties are that all prefixes of any non-dominated sequence must also be non-dominated, and that if the given CTMC has a finite state set, then there must be a finite number of non-dominated state sequences given a total time $T$. Further, listing all the non-dominated sequences and checking which has the highest probability, one is guaranteed to find the maximum probability sequence. Finally, they designed a path extending and pruning algorithm that enumerates the non-dominated sequences from shorter to longer, starting from the specified starting states. After enumerating, each non-dominated sequence with the specified ending state is evaluated to see which is the most probable at time $T$.  This method does not have an easy complexity analysis but it is proved to terminate in finite time. They named the method \textit{State Sequence Analysis (SSA)}. Later work, \cite{Grinberg2015}, has also been able to show SSA as an better inference method for decoding in DT-HMM scenarios compared to MAP inference methods, but to the best of our knowledge, our work introduces the first and only approach towards decoding CT-HMM state trajectories by combining this SSA methodology with Viterbi decoding. 

\subsection{State trajectory decoding in CT-HMM}
\label{sec:state-trajectory-ct-hmm}
Below we describe our state trajectory decoding method for CT-HMM.  This leverages both the \textit{SSA} method for CTMC decoding and recent advances in computing the expected state dwelling time. Our proposed decoding method has the following three properties. First, the decoded states at the observation times are optimal if considering all possible internal paths and state dwelling times. Second, the decoded inner state transition path in between two successive decoded states at the observation times are optimal if considering all possible state dwelling time in the time interval. Third, the expected state dwelling time is assigned to the decoded inner state transition path given the time interval. This dwelling time assignment may not be the maximum likelihood one but can be more representive for the system behavior (note that the maximum likelihood dwelling time assignment will assign all the time to the state with the smallest $q_i$ (\cite{Perkins2009NIPS}).

\textbf{
First, decode the optimal state sequence \textit{at} the observation times.}
Given the observed data $O=(o_0, o_1,\ldots, o_{V})$ at time $T=(t_0, t_1,\ldots,t_{V})$ and the known state transition  matrix $Q$, the goal here is to find the best state sequence $S^*_O=(s(t_0), s(t_1),..., s(t_V))$ at the observation times (we also refer it as the \textit{outer} state sequence), which can be  formulated as:
{
\begin{align}
S^*_{O} = arg~ max_{S_O} ~~p(S_O|O,T,Q) \label{eq:decoding_step1}
\end{align}
}
The key is to note that the posterior state probabilities are only needed at the observation times. We can exploit this insight to reformulate the decoding problem here in terms of an equivalent discrete time-\emph{inhomogeneous} hidden Markov model (similar idea as in the learning section \ref{sec:computing-posterior-state}).
Specifically,  we divide the time into $V$ intervals, each with duration $\tau_v = t_v - t_{v-1}$. We then make use of the transition property of CTMC, and associate each interval $v$ with a state transition matrix $P^v(\tau_v):=e^{Q \tau_v}$. Together with the state emission model $p(o|s)$, we then have a discrete time-inhomogeneous HMM with joint likelihood for a state sequence $S_O = (s(t_0), s(t_1),..., s(t_V))$ as:
{
	\begin{align}
	p(S_O|O,T,Q) = \prod_{v=1}^V [P^v(\tau_v)]_{(s(t_{v-1}), s(t_v))} \prod_{v=0}^V p(o_v | s(t_v)).\label{eq:inHMM_decoding}
	\end{align}
}%
This formulation allows us to reduce the computation of
$S^*_{O}$ to familiar operations in discrete-time HMM. The MAP assignment of hidden states can thus be obtained using the \textit{Viterbi} algorithm.

\textbf{
Second, decode the best state sequence for each end-state and time interval conditioned segment.}
after finding the best states at observation times, we then compute the best (inner) state sequence $S^*_{v}=(S^*_{O}(v),..., S^*_{O}(v+1)) $ for every time segment  $\tau_v = t_{v+1} - t_v$ between adjacent visits $v$, $v=1,...,V$, where the best end-states $(S^*_{O}(v)$ and $S^*_{O}(v+1))$ have been calculated from the first step. This is formulated as:
{
\begin{align}
 S^*_{v} = arg~ max_{S_v} ~~p(S_v|S^*_{O}(v), S^*_{O}(v+1),\tau_v, Q) 
\end{align}
}
To compute the best inner state sequence between two given end-states with a total time interval in this CTMC, we apply the \textit{SSA} algorithm from \cite{Levin2012}.

\textbf{
Third, calculate the expected state dwelling time for the best (inner) state sequence for each time segment.}
We compute the expected state dwelling time for each state in the best (inner) state sequence $S^*_{v} = (s_{1}, s_{2}, ..., s_{m})$ for $v=1,...,V$ from the previous step. 
To achieve this, for each $v$, we can construct an auxiliary transition rate matrix $Q_v$ with dimension $(m+1) \times (m+1)$ using the state sequence $S^*_{v}$ as follows:
\begin{align}
Q_v = \begin{bmatrix}
-q_{s_1} & q_{s_1}  & 0            & \cdots &  0  & 0  \\
0             & -q_{s_2} & q_{s_2} & \cdots &  0  & 0  \\    
\cdots & \cdots & \cdots & \cdots & \cdots & \cdots \\
0 & 0 & 0 & \cdots & -q_{s_m} & q_{s_m} \\
0 & 0 & 0 & \cdots & 0 & 0 \\
\end{bmatrix}_{(m+1)\times(m+1)}
\end{align}
where $q_{s_k}$ is the holding time parameter of state $s_k$. This matrix has the structure that only the transition from $s_i$ to $s_{i+1}$ is set positive at the $(i,i+1)$ entry, and thus the state transition must follow the specified order in $S^*_{v}$. 
The same construction of $Q_v$ is also used in \cite{Hajiaghayi2014} (Eqn. 4,5) for computing the path-and-time conditioned state sequence probability, while it is utilized here for expected dwelling time calculation.


Now, to compute the expected state dwelling time $\tau^*_{v,i}$ for each state $s_i$, in the state path $S^*_{v}$ with a total time $\tau_v$, we can
simply apply the same \textit{Expm, Unif}, or \textit{Eigen} method to compute the \textit{end-state} conditioned statistics with the $Q_v$ matrix as in the learning section \ref{sec:em}, as the only allowed path has been explicitly specified in $Q_v$:
\begin{align}
	 \tau^*_{v,i} = \mathbb{E}[\tau_{S^*_{v}(i)}| S^*_{v}, \tau_v, Q_v], 1 \leq i \leq m, 1 \leq v \leq V
\end{align}
Specifically, this implies that $p(s(t_{v})=k,s(t_{v+1})=l|O,T,\hat{Q}_{0})=1$ for the given end-states associated with each (inner) state sequence and $0$ elsewhere.


\textbf{
	Finally, combine the results from the above, and generate an overall state trajectory $\{ (s^*_1, \tau^*_1), ..., (s^*_k, \tau^*_k) \}$ that spans the entire time $T$ (i.e., $\sum_{i=1}^k \tau^*_i = T)$ .}

Our decoding algorithm for CT-HMM, dubbed the name \textbf{\textit{Viterbi-SSAE}}, where \textit{SSAE} represents the SSA algorithm and the expected dwelling time calculation, is detailed in Algorithm \ref{alg:traj_decode}. 
\begin{figure}[!t]
	\centering
	\includegraphics[width=12cm]{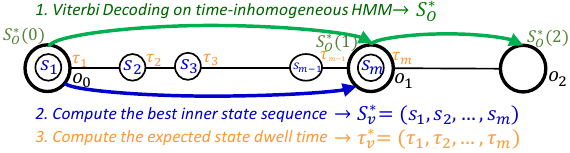}
	\caption{
Illustration of the CT-HMM state trajectory decoding method \textit{Viterbi-SSAE}.
	}
	\label{fig:CTHMM_decode_algo}
\end{figure}


\begin{algorithm}[!ht]
	\caption{CT-HMM State Trajectory Decoding: Viterbi-SSAE}
	\label{alg:traj_decode}
	\begin{algorithmic}[1]
		\STATE {\bfseries Input:} data $O=(o_0, ..., o_V)$, observation time $T=(t_0,\ldots, t_V)$, states $S$, transition rate matrix $Q$ 
		\STATE {\bfseries Output:} decoded state sequence and corresponding dwelling times  $\{ (s^*_1, \tau^*_1), ..., (s^*_k, \tau^*_k) \}$		
		\STATE Compute the best states at observation times $S_{O}^* = max_{S_{O}}~~p(S_{O}|O,T,Q)$ by constructing an equivalent time-inhomegeneous HMM and using Viterbi decoding
		\FOR{$v=1$ {\bfseries to} $V-1$}
		\STATE Set time interval $\tau_v = t_{v} - t_{v-1}$
		\STATE Compute the best inner state sequence $S_{v}^* = max_{S_{v}}~~p(S_{v}|S_{O}^*(v), S_{O}^*(v+1), \tau_v, Q)$ using \textit{state sequence analysis (SSA)} algorithm from \cite{Levin2012}
		\STATE Compute the expected state dwelling time for each state in $S_{v}^*$ using 
		\textit{Expm, Unif} or \textit{Eigen} method on the auxiliary transition rate matrix $Q_v$ constructed from $S_{v}^*$
		\STATE Combine the decoded results so far
		\ENDFOR	
	\end{algorithmic}
\end{algorithm}

%





	
 
\textbf{\\ Time complexity analysis for calculating the expected state dwelling time (algorithm step 7):}

Table \ref{table:compare_time_expect_duration} shows the complexity analysis for evaluating the path-and-time conditioned expected state durations.  As in EM learning, the \textit{Eigen} method has the best complexity at $O(m^3)$ $O(m^3)$ to calculate the path-and-time conditioned expected state durations, where $m$ is the number of states in the path, however, it requires $Q_v$ to be diagnalizable, while \textit{Expm} and \textit{Unif} method has no such constraints. 

\begin{table*}[!ht]

		\centering	
		\newcommand{\tabincell}[2]{\begin{tabular}{@{}#1@{}}#2\end{tabular}}
		\caption{
			Time complexity comparison of all methods in evaluating path-and-time conditioned expected state durations.
			(m: number of states in the path, $M$: the truncation point for \textit{Unif} for $Q_v$, set as $\ulcorner 4 + 6 \sqrt{\hat{q_v} t} + (\hat{q_v} t) \urcorner$, where $\hat{q_v} = max_i ~q_i$ for states in the path, and $t$: the total time).
		}	
		\label{table:compare_time_expect_duration}
		\begin{tabular}{|l||l|l|l|}
			\hline
			Complexity & Expm & Unif & Eigen \\ \hline \hline
			For one state & $O(m^3)$ & $O(M m^3 + m^2)$ & $O(m^3)$\\ \hline
			For all states & $O(m^4)$ & $O(M m^3 + M^2 m)$ & $O(m^3)$ \\ \hline \hline
			Prerequisite & none & none & Q diagonalizable and well-conditioned \\ \hline		
		\end{tabular}

\end{table*}


\subsection{Experimental Results}
\label{sec:exp-decode}

\subsubsection{Simulation on a Toy Example}

\begin{figure}[!ht]
	\centering
	\includegraphics[width=4.5cm]{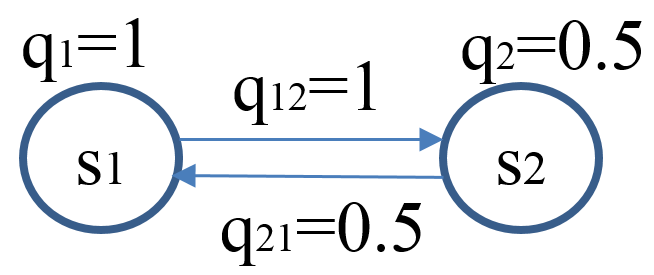}
	\caption{
		A 2-state digraph used to demonstrate the optimal state trajectory decoding. 
	}
	\label{fig:2_state_digraph}
\end{figure}

We now show one toy simulation example, where we compute the optimal state trajectory given two known end-states and a total time using our decoding algorithm (step 6 and 7). 
In Fig.~\ref{fig:2_state_digraph}, we show a simple 2-state complete digraph models, with $q_1=1$ (mean holding time = 1) and $q_2=0.5$ (mean holding time = 2). If we set the beginning state to be $s_1$, the terminating state to be $s_2$ and a total duration = $12$, we would like to compute the optimal state trajectory.  

Using algorithm step 6, the found best state sequence considering all possible dwelling time assignments is
$(1,2,1,2,1,2,1,2)$.
Then, step 7 of our method computes the expected state duration for each state given the found best state sequence and a total time 12, which derives the following duration sequence: $(1,2,1,2,1,2,1,2)$.
This simulation result corresponds to our intuition for the best state path and the corresponding expected dwelling time for this given model, where state $1$ has average  holding time $1$, and state $2$ has average holding time $2$.

\subsubsection{Trajectory Decoding Simulation on a 5-state Complete Digraph}

We test the accuracy of our decoding algorithm on a 5-state complete digraph with synthetic data generated under different observation intervals and noise levels.
The ground truth $Q$ matrix is set up such that 
each $q_i$ is randomly drawn from $[1,5]$ and then $q_{ij}$ is drawn from $[0,1]$ and renormalized such that $  \sum_{j \neq i} q_{ij} = q_i$. The state chains are generated from $Q$, such that each chain has a total duration around $T = \frac{100}{\min_i q_i} $, where $\frac{1}{\min_i q_i}$ is the largest mean holding time among the states. The data emission model for state $i$ is set as $N(i, \sigma^2)$, where $\sigma$ varies under different noise level settings (the larger the $\sigma$ the higher the noise.)
The observations are then sampled from the state chains with time intervals $\tau = \tau_s \times {\max_i q_i}$, where $\frac{1}{\max_i q_i}$ is the smallest mean holding time, and the sampling interval parameter $\tau_s = 0.5, 1, 2, $ and $4$.
A total of $10^4$ observations are sampled from state chains for each setting. 

We calculate both the continuous-time and observation-time decoding error. For the former, we compute the error rate measuring the difference between the decoded state trajectory and the original state chain in continuous-time. For the later, we compute the difference between states only at the discrete observation times using the results of \textit{Viterbi} decoding (from algorithm step 3). The results from 5 random runs for each setting are listed in Table \ref{table:decode_accuracy}\text{.}

From Table \ref{table:decode_accuracy}, we can see that the denser the observation interval is, the closer the decoded most probable state trajectory is to the underlying true state chain for both continuous-time and observation-time decoding error. This makes sense because when the observation interval is shorter, more evidence is provided in the same period of time, and thus the decoding results are less ambiguous with respect to the true state sequence. 
Note that a random guess is 80$\%$ error rate for this 5-state model for both the continuous and discrete-time error metrics. 

In table \ref{table:decode_time}, we show the average running time for computing one optimal inner best state sequence (\textit{SSA} method from \cite{Levin2012}, code is provided) and its corresponding state dwelling time (\textit{Expm} is used here) 
in the 5-state simulation, on a 1.8GHz machine using Matlab. 






\begin{table*}[!htb]
	\centering	
	\newcommand{\tabincell}[2]{\begin{tabular}{@{}#1@{}}#2\end{tabular}}
	\caption{
		Continuous-time decoding error (the first line in each cell) and observation-time decoding error (the second line in each cell) under different observation time intervals ($\tau_s \times {\max_i q_i}$), and noise levels $\sigma$ in the observation model $N(i, \sigma^2)$.
	}	
	\begin{tabular}{|l||l|l|l|l|}
		\hline
		Error & $\tau_s=0.5$ & $\tau_s=1.0$ & $\tau_s=2$ & $\tau_s=4$\\ \hline \hline
		$\sigma = 1/4$ & \specialcell{0.0674 $\pm$ 0.0089 \\ 0.0107 $\pm$ 0.0021} & \specialcell{0.1623 $\pm$ 0.0426 \\ 0.0235 $\pm$ 0.0044} & \specialcell{0.2454 $\pm$ 0.0333 \\ 0.0285 $\pm$ 0.0040} & \specialcell{0.3967 $\pm$ 0.1103 \\ 0.0327 $\pm$ 0.0033} \\ \hline
		$\sigma = 1/2$ & \specialcell{0.1704 $\pm$ 0.0365 \\ 0.1215 $\pm$ 0.0243} & \specialcell{0.2509 $\pm$ 0.0170 \\ 0.1723 $\pm$ 0.0072} & \specialcell{0.3081 $\pm$ 0.0699 \\ 0.1972 $\pm$ 0.0391} & \specialcell{0.4710 $\pm$ 0.0529 \\ 0.2145 $\pm$ 0.0061}  \\ \hline
	\end{tabular}	 
	\label{table:decode_accuracy}
\end{table*}


\begin{table*}[!htb]
	\centering	
	\newcommand{\tabincell}[2]{\begin{tabular}{@{}#1@{}}#2\end{tabular}}
	\caption{	
		Average running time for computing one optimal inner best state sequence (\textit{SSA} method from \cite{Levin2012}) and its corresponding state dwelling time (\textit{Expm} is used here) given one pair of end-states and a total time ($\tau_s \times {\max_i q_i}$) in the 5-state simulation.
	}	
	\begin{tabular}{|l||l|l|l|l|}
		\hline
		Run-Time & $\tau_s=0.5$ & $\tau_s=1.0$ & $\tau_s=2$ & $\tau_s=4$\\ \hline \hline
		SSA (step 6)& \specialcell{5.0881 sec} & \specialcell{5.3491 sec} & \specialcell{5.9852 sec } & \specialcell{6.3929 sec} \\ \hline	
		
		Expm (step 7)& \specialcell{0.0007 sec} & \specialcell{0.0007 sec} & \specialcell{0.0007 sec} & \specialcell{0.0007 sec} \\ \hline	
	\end{tabular}	 
	\label{table:decode_time}
\end{table*}


\subsubsection{Visualization of the Decoded Trajectory on Glaucoma Dataset}
\label{sec:glauc}

\begin{figure}[tb]
	\vskip -0.1in
	\centering	
	\subfloat[]{\includegraphics[width=7cm]{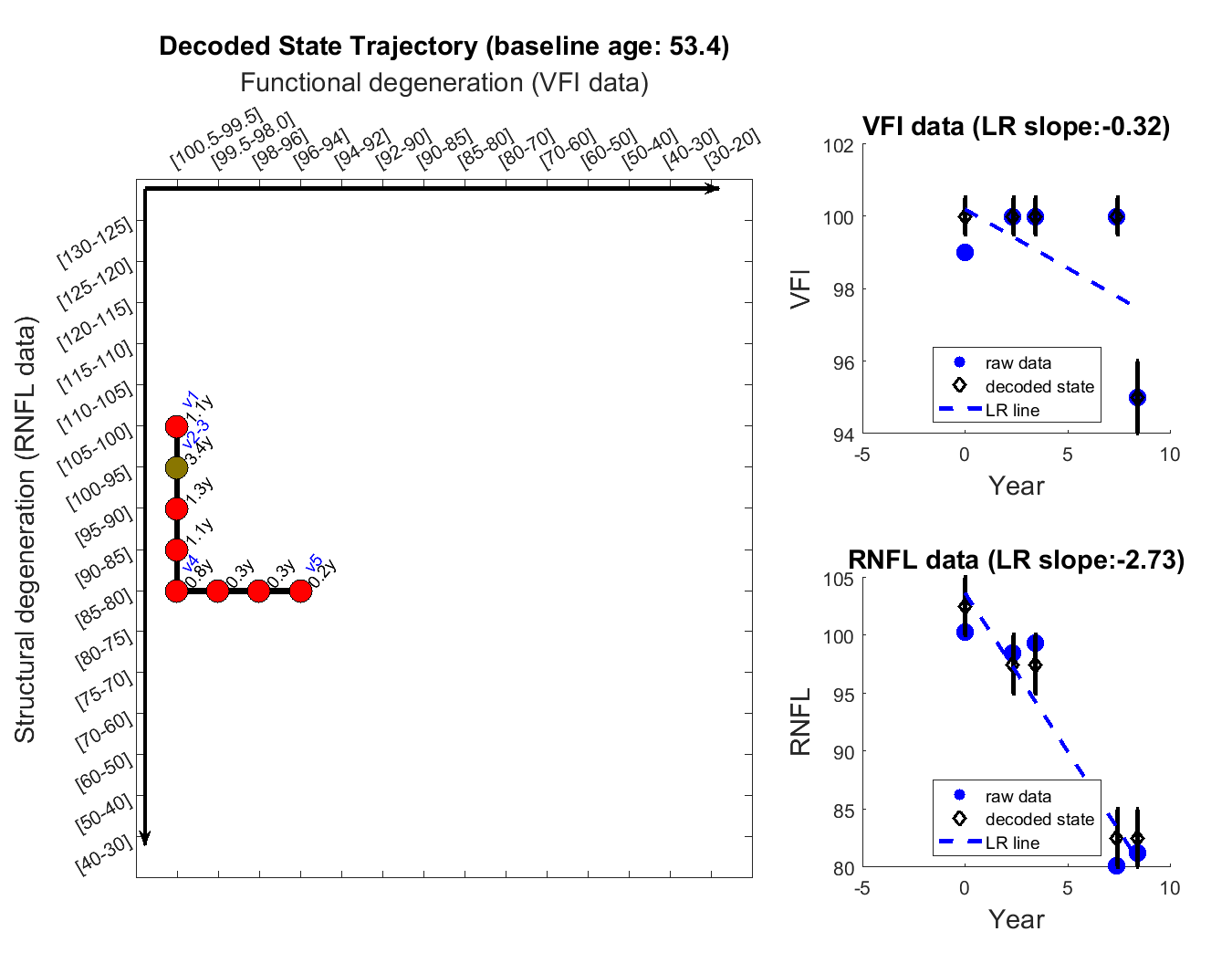}}
	\subfloat[]{	\includegraphics[width=7cm]{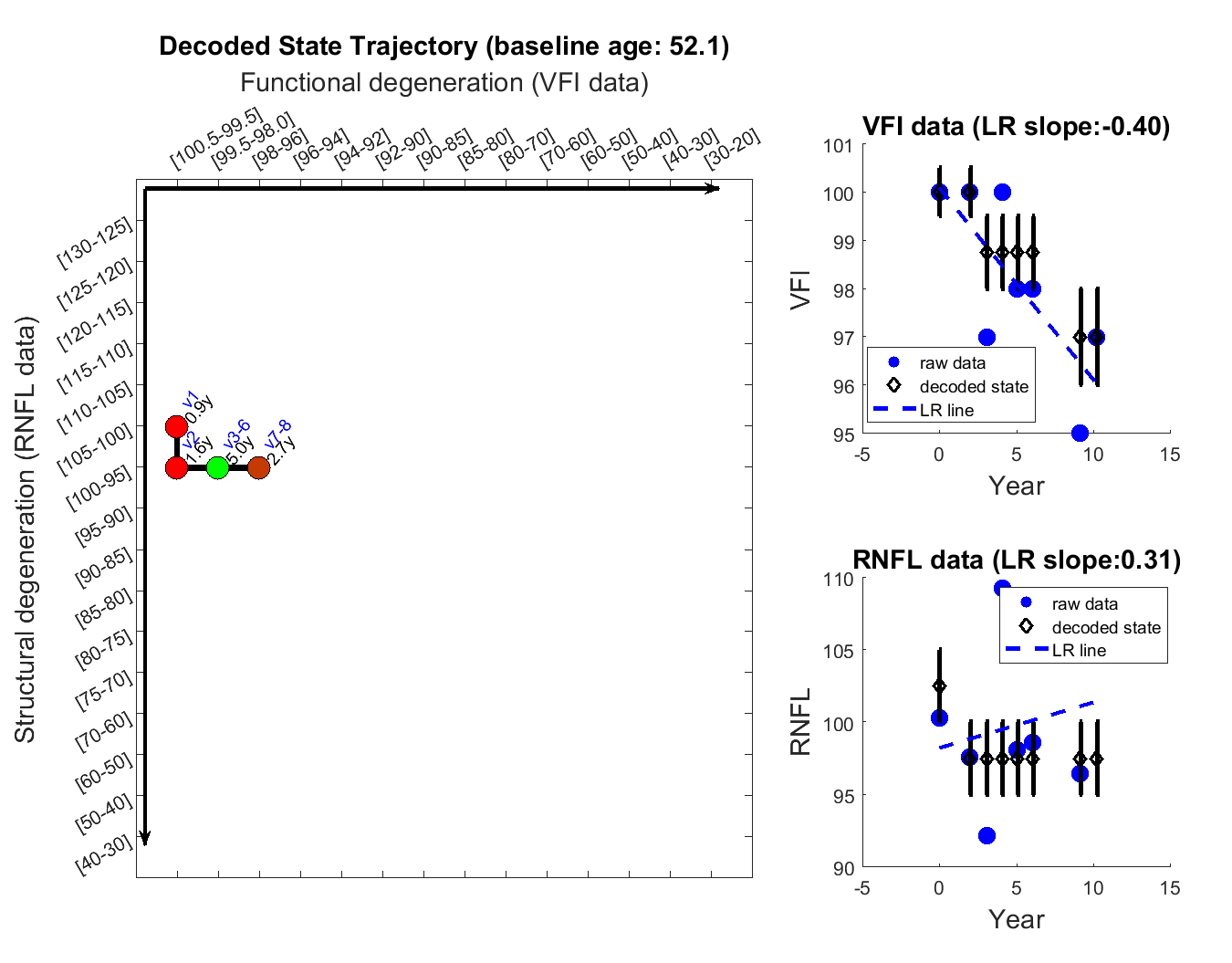}}\\
	\subfloat[]{\includegraphics[width=7cm]{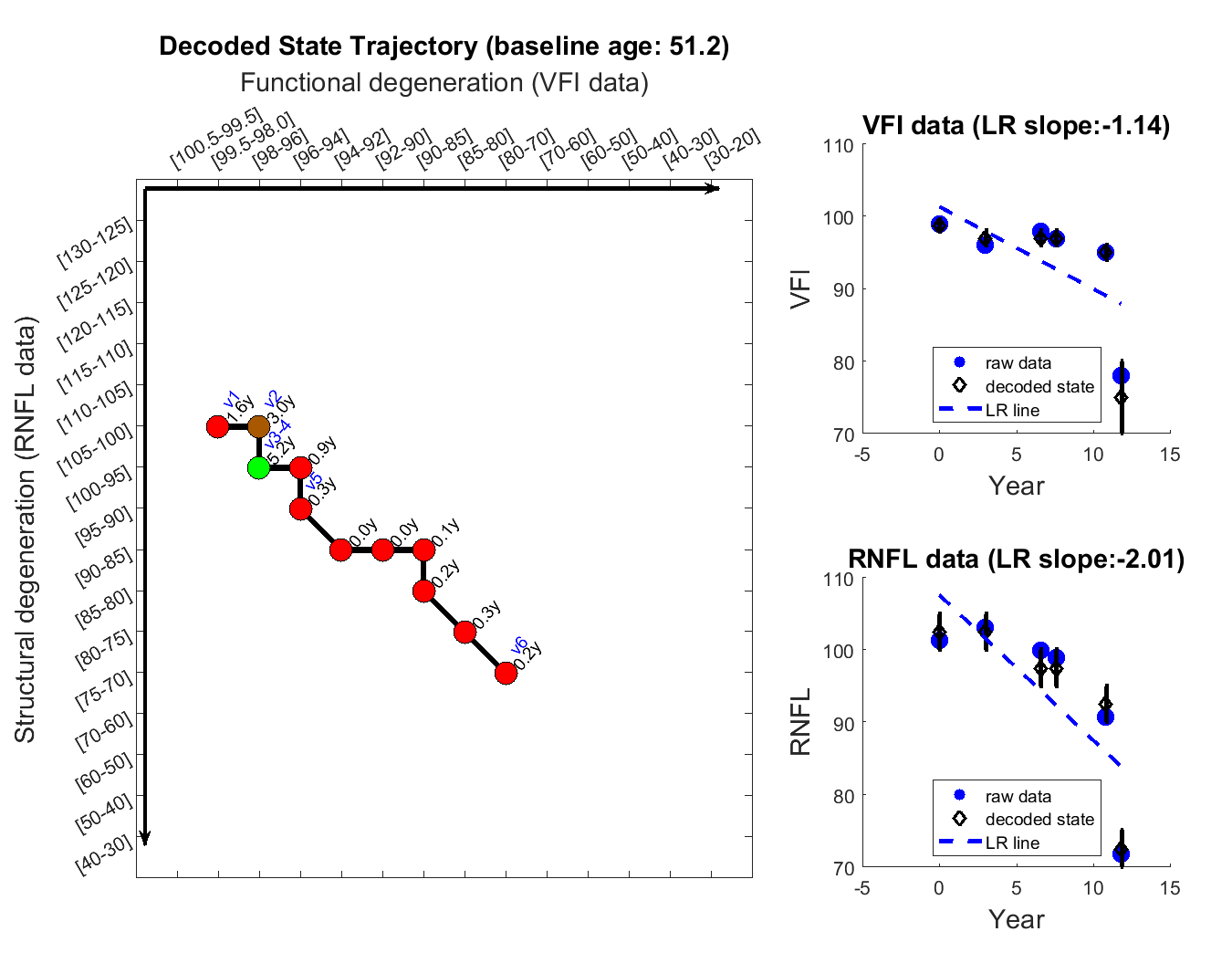}}
	\subfloat[]{\includegraphics[width=7cm]{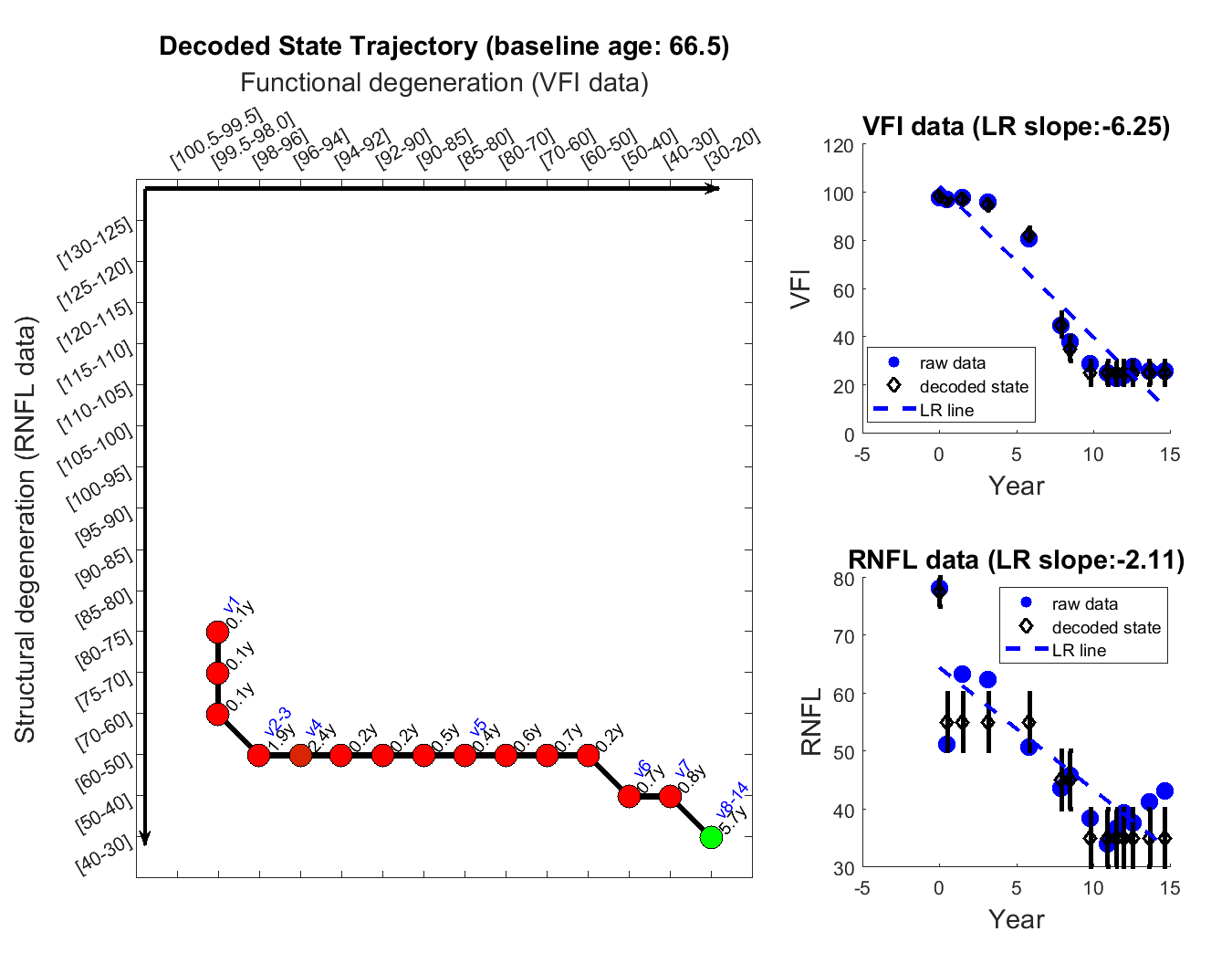}}
	\vskip -0.05in
	\caption{	\small
The continuous state transition path (left side) and the longitudinal raw data (right side, VFI: visual field index, and RNFL: retinal nerve fiber layer, LR: linear regression line) from four glaucoma patients. In the left side, the node color represents the decoded state dwelling time (red: $<$ 2 years, red to green: 2 to 5 years and above). 
In the right side, blue dots and dashed line represents the raw data and its linear regression respectively, and the black diamond shows the decoded CT-HMM state for the raw data.
	}
	\label{fig:glaucoma_decode}
	\vskip -0.15in
\end{figure}

By using the decoding method, we can decode and visualize the most probable continuous state transition trajectory for each patient in an intuitive way. Here, we show several decoded examples from our glaucoma dataset (described in Section \ref{sec:exp2}) in Fig. \ref{fig:glaucoma_decode}, where each patient's data is decoded using the model trained from all the data. 

In Fig. \ref{fig:glaucoma_decode}, the continuous trajectory in the state space is shown for each example where the state color represents the state dwelling time and the line between nodes denotes the state transition path. 
From Fig. \ref{fig:glaucoma_decode}(a), we can clearly see that the patient maintains good visual field (VFI=100) while having loss of retinal nerve fibers (decreasing thickness of retinal nerve fiber layer (RNFL)), and until the thickness of RNFL reachs that state (85-80), visial field starts to degenerate. 
	This example shows one instance that the state  (VFI=100, RNFL=[85-80]) is an important transition point from structural to functional degeneration (see Fig.  \ref{fig:glaucoma}(a), the state transition trend visualization).
From Fig. \ref{fig:glaucoma_decode}(b),  we show another example where a higher level of RNFL thickness (VFI=100, RNFL=[100-95]) is the transition point from structural to the begining of functional loss. Checking from the global model visualization in Fig.  5(a), we can also find the strong transition from this state to functional loss, though continuing on only structural loss is the most probable one.

In Fig. \ref{fig:glaucoma_decode}(c)(d), two patients with fast progressing trajectory can be easily perceived from the red color of states (short state duration). In Fig. \ref{fig:glaucoma_decode}(c), functional and structural degeneration seems to happen together, but in Fig. \ref{fig:glaucoma_decode}(d), visual function remains good until the retinal fiber layer becomes very thin, after which there is rapid functional degradation. These two examples may represent two distinct phenotypes of glaucoma progression.
	
Besides visualizing a patient's state transition path, one may also discover patients of similar states/state paths by direct grouping or through a clustering method with a similarity comparison metric. Then, training and visualizing each patient group can help gain insights of possibly progressing phenotypes for further investigation by the experts. For prediction purpose, one may also find the nearest patients by using history path matching, and using their future trajectory for reference in clinics. 

By using the proposed decoding method with the intuitive visualization, a patient's longitudinal information as well as the transition characteristics from a cluster of patients, which are found through using a trajectory clustering algorithms, can be easily comprehended in a global view, which may be valuable in clinics.

\subsubsection{Visualizing Language Development of Children with ASD with Decoded Latent State Trajectories}

Many children with autism spectrum disorder (ASD) will develop some speech and language skills, but not to the level of ability found in typical development. Typically developing children produce their first words by 12 months of age and then rapidly acquire new words, reaching a 200\textendash300 word vocabulary by 24 months and 900\textendash1000 words by 3 years of age (\cite{loraine2008vocabulary}). In contrast, children with autism experience significant delays in language acquisition, with as many as a 30\% producing very few if any spoken words by the time they reach kindergarten (\cite{ASDVerb}). The development of early predictive markers that can identify those children with ASD who will acquire language and those who will remain delayed or minimally-verbal is a key area of research, with implications for treatment. One such hypothesized early risk factor is the presence of specific developmental inflection points during which children’s language development may become stalled, precluding progression into full spoken language use \cite{InflectionPoint2013}. The goal of this section is to explore the utility of the state trajectory decoding for CT-HMMs to model the time course of language acquisition and identifying potential developmental inflection points.

As in the case of glaucoma patients in Sec.~\ref{sec:glauc}, the timing of the clinical visits of children with ASD will be variable,  resulting in a time-series with irregularly sampled observation points. This property, combined with the fact that language development is a continous process, makes the CT-HMM model and proposed Viterbi-SSAE decoding methodology an attractive approach for describing trajectories of development and addressing the inflection point hypothesis. Each child's latent state trajectory can model their latent language development and capture information about current and past progression. For the purposes of this analysis, we utilized data from a longitudinal study by \cite{Yoder2015} consisting of 87 children with ASD (mean age at study intake of 35.2 months, range 20-58 months) who were minimally verbal at intake, and were then observed across 5 time points spanning 16 months. The data consisted of longitudinal measurements of parent-reported word production (total words said on the MacArthur Communicative Development Inventory; MCDIPV) (\cite{Hutchins2013}) and clinician-documented spoken word use during an observational assessment which was performed during clinic visits (weighted raw score for words on the Communication and Symbolic Behavior Scales; CSBSWDS) (\cite{Morris2013}). Figure 10 below shows the trajectories of the 87 children across these two measures. For reference, an increase in either measure corresponds to an increase in language development.

\begin{figure}[!ht]
	\centering	
	\includegraphics[width=5cm]{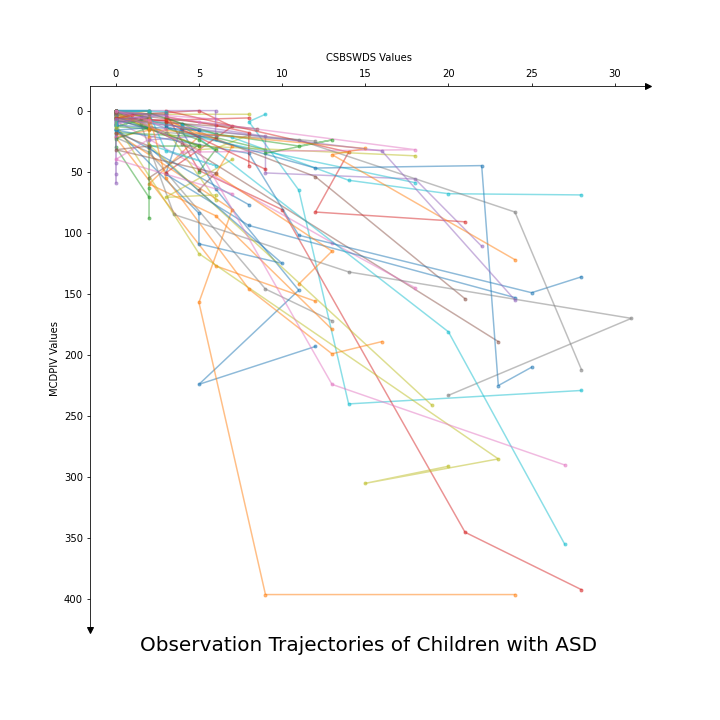}
	\vskip -.05in
	\caption{	\small
    Observation trajectories of 87 children across the MCDIPV and CSBSWDS language development metrics.
	}
	\label{fig:asd_obs_trajs}
\end{figure}

We performed three tasks in evaluating the CT-HMM's potential as a modeling tool for verbal and nonverbal development in ASD. First, we jointly calibrated the the MCDIPV and CSBSWDS measures so that they are aligned in measuring developmenal progress. Next, we apply the Viterbi-SSAE decoding method to compute each child's state transition path and then visualize them to examine and characterize the differences between the trajectories of the verbal and nonverbal subpopulations within our ASD cohort. Third, we utilized these visualizations to potentially identify states in which development is stalled, thus helping guide the determination of the timing of interventions.

Similar to the glaucoma model, each state's emission model was set up to be centered on a 2D-grid. Each emission model was a bivariate normal distributions with diagonal covariances of the MCDIPV and CSBSWDS metric. The transition matrix was constrained as a 2D version of the standard left-to-right model (i.e. no backward transitions are allowed) with a neighborhood structure without skip-state transitions (i.e. states are not fully connected: in order to progress through the states, each subject must proceed to the state immediately below and/or to the right of its current state). The left-to-right constraint eliminates the possibility of regression, which is developmentally rare but can arise in developmental data due to measurement error and noise. Likewise, the use of a model without skip-state transitions reflects the fact that progression through developmental milestones should be monotonic with respect to our developmental measures, even though language development may take place at different rates over time. In Fig. 11a, we can visualize the decoded state trajectories of the patients learned by the CT-HMM model.

\begin{figure}[!ht]
	\vskip -0.1in
	\centering	
	\subfloat[]{\includegraphics[width=5cm]{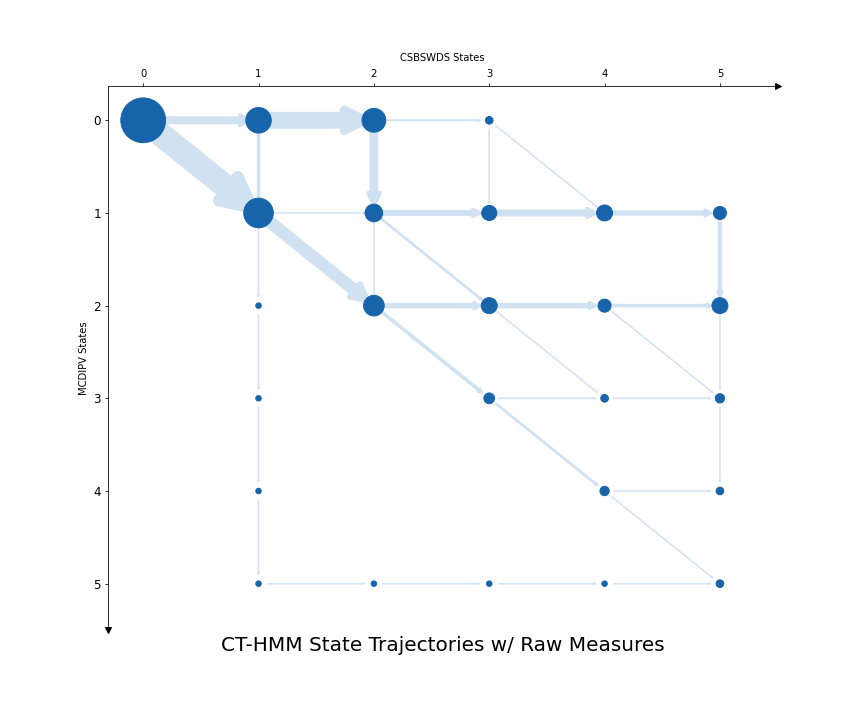}}
	\subfloat[]{\includegraphics[width=5cm]{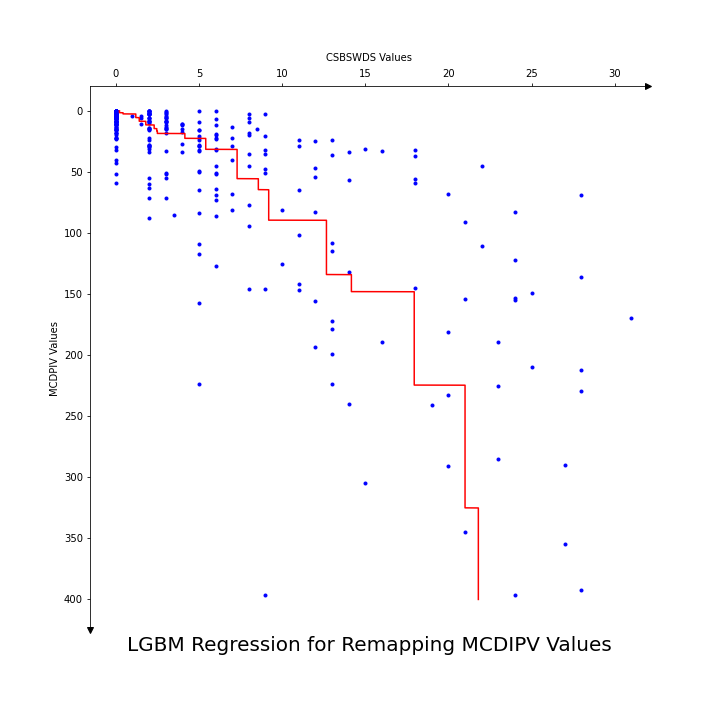}}
	\subfloat[]{\includegraphics[width=5cm]{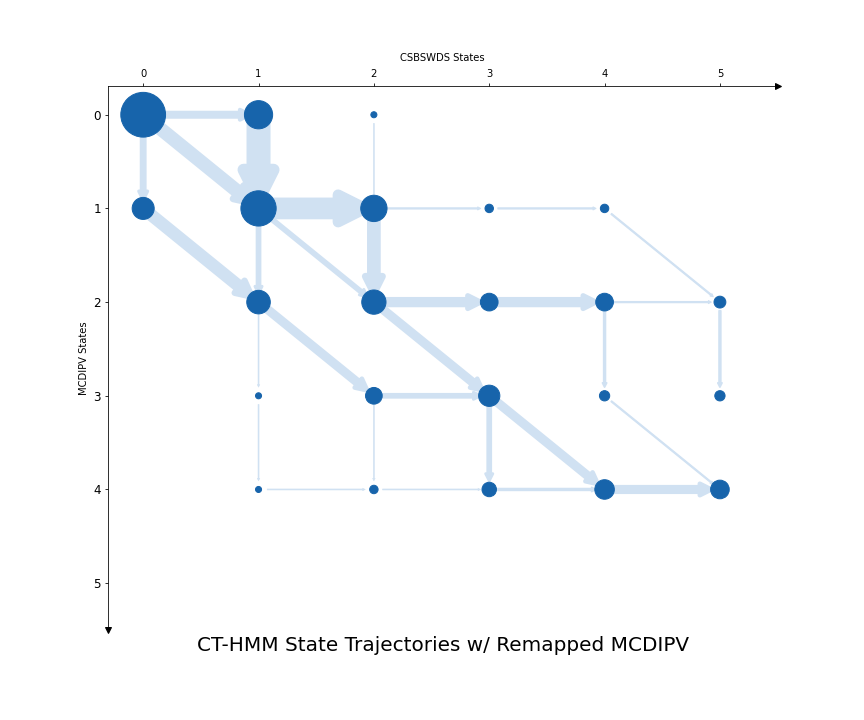}}
	\vskip -0.05in
	\caption{	\small
a) State trajectories of children with ASD using a model trained on raw CSBSWDS and raw MCDIPV values. Size of the nodes and arrows correspond to the number of children who entered that state or the number of children that made that specific state-to-state transition respectively. b) LGBM regression of raw MCDIPV values to match the same scaling of the CSBSWDS values as illustrated by the red line. The blue dots are a scatter plot of the original raw MCDIPV values versus raw CSBSWDS values. c) State trajectories of childen with ASD with model trained on aligned CSBSWDS and MCDIPV scales.
	}
	\vskip -0.15in
\end{figure}

Our first task was to address the joint calibration of the MCDIPV and CSBSWDS measures. The spaghetti plot in Fig. 10 demonstrates that while CSBSWDS and MCDIPV both measure language development, progression appears to be faster along the CSBSWDS dimension. This same pattern can be observed in the CT-HMM state progression based on the raw scores, as illustrated in Fig. 11a. In order to define a set of latent developmental states that are informed by both measures, it is necessary to address their misalignment (which results from differences in scaling due to a lack of joint calibration). 

To address this issue, We utilized a Light Gradient Boosting Machine Regression model to align the scaling of MCDIPV to the CSBSWDS with a monotonicity constraint \cite{LGBM}. The regression plot is illustrated in Fig. 11b. We then used the remapped MCDIPV observation values combined with the original CSBSWDS measures to train a CT-HMM model, resulting in the decoded state trajectories shown in Fig. 11c. In comparison to Fig. 11a, the state trajectories are clustered along the main diagonal (as evidenced by the presence of states and transitions with a high count). This more accurately reflects the fact that as a child's vocabulary develops, their scores on both CSBWDS and MCDIPV should both improve in tandem. 

Having calibrated our two measures of language development, the next step is to examine the Viterbi-SSAE decoded state transitions to identify any patterns that reflect known characteristics of language development. We observe that 45 out of 87 children made minimal progress on language development, being unable to progress past state 1 on either the MCDIPV state or CSBSWDS state. Out of these 45, we find that 30 of these children remained in the initial (0,0) state \footnote{For the state pairing, the first index corresponds to the CSBSWDS state and the second index corresponds to the MCDIPV state}. In order to examine differences in verbal development in more detail, we stratified our sample into two subpopulations based on well-established criteria for defining nonverbal outcomes. The two disjoint subpopulations, \emph{Minimally Verbal} (n=45) and \emph{Verbally Developed} (n=33) were based on a clinical ascertainment of the language sample gathered at the final visit. Nine children were excluded from this analysis due to missing the final visit. 

Fig. 12a shows the state trajectories of the verbally developed children. We found that 25 out of the 33 children achieved signficant language development by having state trajectories that progressed past the upper left-hand quadrant bounded by (2,2) by the time the 16-month study ended. In contrast, 44 out of 45 of the minimally verbal children in Fig. 12b made relatively little progress, and remained within this quadrant. 

Our last analysis uses the visualizations of Fig. 12 to explore the possibility of defining inflection point states. These inflection points states capture the intution that progress in language development is not uniform in time, and certain states may require more time to exit than others. Children with nonverbal outcomes in particular may experience substantial difficulties in exiting such states. 
Therefore, inflection point states may act as a candidate criteria for determining the risk for not progressing to spoken language. Analogous to identifying fast progression in glaucoma, we can use CT-HMMs to monitor an abnormally slow progression of language development in minimally verbal children to identify inflection point states. For example, in Fig. 12, we have found (1,1) and (1,2) to possible be indicative of an inflection point risk state. Fig. 12a shows that although there are many verbally developed children who will dwell in the aforementioned state pairs for extended periods of time, once they have passed it, they will quickly develop their language skills. Fig. 12b shows us that most of the minimally verbal children were unable to pass one or the other of the two state pairs and therefore reached a plateau in their language development. 

One potential clinical use for inflection point-derived risk markers would be in identifying the appropriate timing for interventions. All children with ASD will be receiving regular treatment, but if it is determined that a child is not making adequate progress, that would be the basis for increasing the frequency or intensity of therapy, or even changing intervention methods. For example, we observed that the average dwell time for verbally developed children in state (1,1) was 4.69 months, whereras the average dwelling time for minimally verbal children was 8.08 months. It follows that children who remain in this state beyond 4.69 months are exhibiting below average progress, which could be the basis for making changes to treatment. While substantial additional work would be needed before a CT-HMM model could be used in clincal settings to inform treatment, these examples suggest the potential benefits of the approach.

The Viterbi-SSAE decoding methodology allows us to elucidate the language development progression of children with ASD, identifying key states that lead to stalling in the acquiring vocabulary, which may enable a more targeted intervention plan for these children.

\begin{figure}[!ht]
	\vskip -0.1in
	\centering	
	\subfloat[]{\includegraphics[width=8cm]{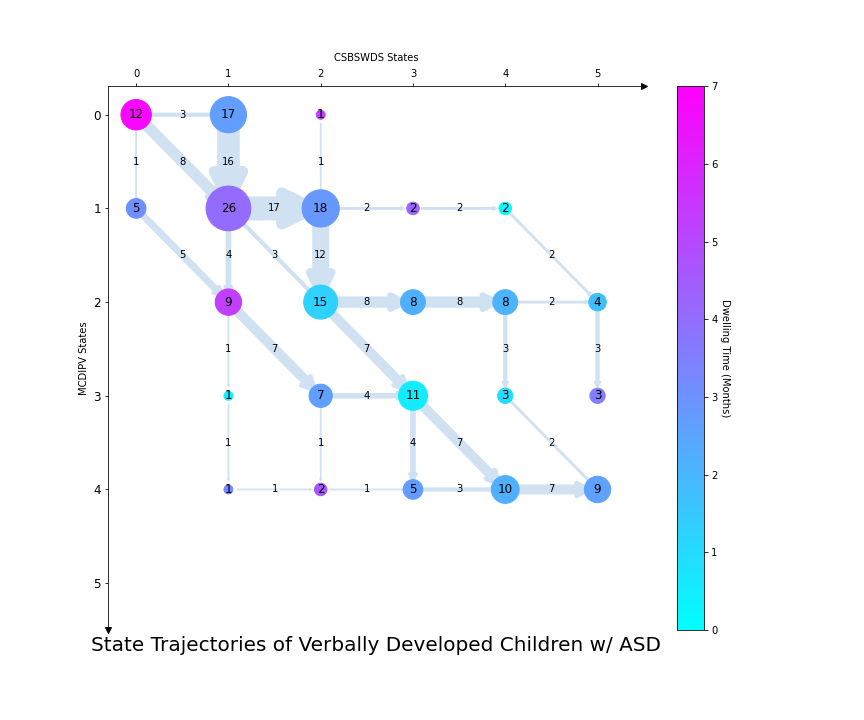}}
	\subfloat[]{\includegraphics[width=8cm]{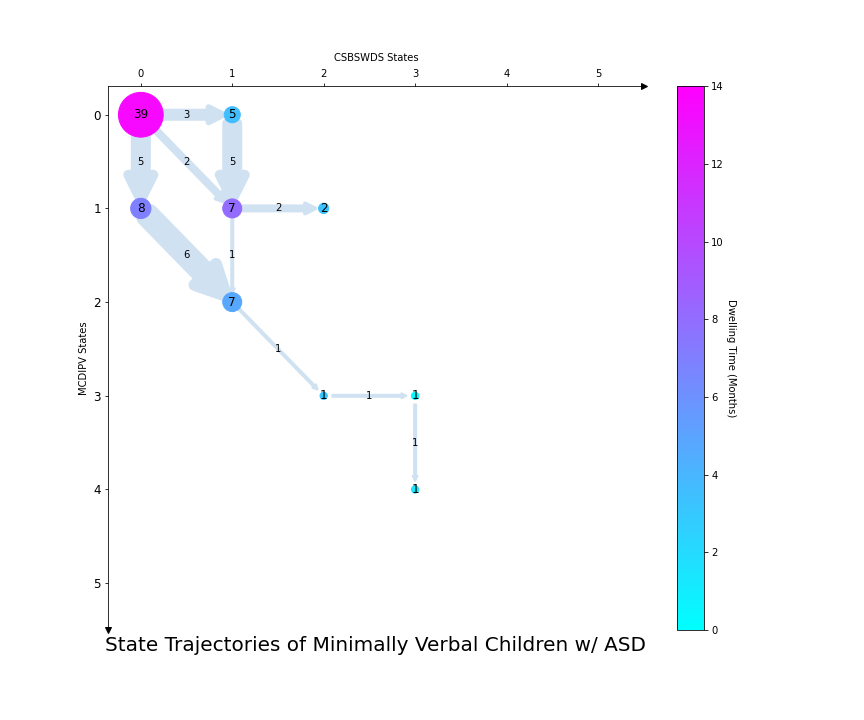}}
	\vskip -0.05in
	\caption{	\small
Size of and number inside the node are the total number of children entered the given state. Color of the node represents the average dwelling time of children who entered this state. Size of and number inside the arrow are the total number of children that make that specific state-to-state transition. a) is the state trajectories of children who were confirmed to have verbally developed by having a sufficient language sample number of different words by the final physician visit and b) is the state trajectories of the children who were not and deemed minimally verbal. 
	}
	\label{fig:statetraj_divided}
	\vskip -0.15in
\end{figure}

\section{Conclusion}
In this paper, we present novel EM algorithms for CT-HMM learning, and the first solution to the optimal state transition trajectory decoding problem in CT-HMM.  For the first problem, we leverage recent approaches from~\cite{Hobolth2011} for evaluating the end-state conditioned expectations in CTMC models.  We improve upon the efficiency of the \emph{soft Eigen} method from our prior work in \cite{LiuNips15} by a factor of number of states in time complexity, and demonstrate in our experiments a 26 to 35 times speedup over \emph{Expm}, the next fastest soft method. To our knowledge, we are the first to develop and test the \emph{Expm} and \emph{Unif} methods for CT-HMM learning.  We present time complexity analysis for all methods and provide experimental comparisons under both soft and hard EM frameworks. We conclude that \emph{soft Eigen} is the most attractive method overall,  based on its speed and its accuracy as a soft method, unless it suffers from an unstable eigendecomposition.  We did not encounter significant stability issues for the Eigen method in our experiments when initialization is done properly. 
We evaluate our EM algorithms on two disease progression datasets for glaucoma and Alzheimer's Disease, and demonstrated that the CT-HMM can provide a novel tool for visualizing and predicting the progression of these diseases. For optimal state decoding, we present a complete algorithm for finding an optimal state transition trajectory in CT-HMM, which leverages the  work from \cite{Levin2012} for CTMC and recent advances in computing the expected state dwelling time. To the best of our knowledge, we are the first to address the problem of finding the hidden state trajectory for CT-HMM, as the prior work for CT-HMM only finds the best states \textit{at} the observation times.
We show the accuracy of our CT-HMM decoding method using a synthetic dataset, and visualize several examples of the decoded state transition trajectory from the glaucoma and ASD language development datasets. The visualization of the decoded best transition trajectory gives a global view of the disease progression of an individual, which also helps gain insights in identifying different potential phenotypes. One may also retrieve patients' data based on the same passed states or state paths for further meta-data analysis or example-based path prediction. From the demonstrated applications on the real world disease data, we believe that our developed learning and decoding tools for CT-HMM will be a valuable tool for clinical use.  The software implementation of our methods is available from our project repository.~\footnote{https://github.com/rehg-lab/ct-hmm}

In future work, we plan to incorporate covariates to model individualized state transition rates and test on a higher-order Markov-model in CT-HMM to include more history information, which may lead to more accurate future path prediction. In addition, more work could be done to improve the computational efficiency of the Expm and Unif methods. As an example, \cite{al2009computing} describes potentially more efficient ways to compute Expm by noting that the upper right corner of the matrix solution is a Fr\'echet derivative, which has its own Pad´e approximation. It appears that the Hadamard and trace manipulations we introduced for the Eigen method could be applied to this approach as well. Scaling and squaring would cancel much of the benefit, so it would have to be replaced by balancing, which has the same goal of reducing the matrix norm. Additional improvements in efficiency would support the development of large-scale state models.

Portions of this work were supported in part by NIH R01 EY13178-15 and by grant
U54EB020404 awarded by the National Institute of Biomedical Imaging and Bioengineering
through funds provided by the Big Data to Knowledge (BD2K) initiative
(www.bd2k.nih.gov). The research was also supported in part by NSF/NIH
BIGDATA 1R01GM108341, ONR N00014-15-1-2340, NSF IIS-1218749, NSF
CAREER IIS-1350983, and funding from the Georgia Tech Executive Vice President
of Research Office and the Center for Computational Health. Additionally, the collection and sharing of the
Alzheimers data was funded by ADNI under NIH U01 AG024904 and DOD award W81XWH-12-2-0012.










\newpage
\bibliography{CT_HMM_ref}
\newpage
\section*{Appendix A: Derivation of \textit{Vectorized Eigen}}

In \cite{Metzner2007},\cite{MetznerJournal2007}, they state that the na\"{i}ve \textit{Eigen} is equivalent to \textit{Vectorized Eigen} without the derivation, which we provide here.  Let
\begin{align}
\tau_{k,l}^{i,j}(t) 
= \sum_{p=1}^{n} U_{kp} U_{pi}^{-1} \sum_{q=1}^{n} U_{jq} U_{ql}^{-1} \Psi_{pq}(t)
\end{align}
where the symmetric matrix $\Psi(t)=[\Psi_{pq}(t)]_{p,q \in S}$ is defined as:
\begin{align}
\Psi_{pq}(t) = \begin{cases}
t e^{t \lambda_p} \text{~~if~} \lambda_p = \lambda_q\\
\frac{e^{t \lambda_p} - e^{t \lambda_q}}{\lambda_p - \lambda_q} \text{~~if~} \lambda_p \neq \lambda_q
\end{cases}
\end{align}
Again letting $V=U^{-1}$, this is equivalent to
\begin{align}
\tau_{k,l}^{i,j}(t)&=[U[V^{T}_{i}U_{j}\circ \Psi]V]_{kl}
\end{align}
To see why, first, note that for the outer product,
\begin{align}
V_{i}^{T}U_{j}\circ\Psi&=\left(\begin{array}{ccc}
U_{1,i}^{-1}U_{j,1}\Psi_{1,1} & \cdots & U_{1,i}^{-1}U_{j,n}\Psi_{1,n}\\
\vdots & & \vdots \\
U_{n,i}^{-1}U_{j,1}\Psi_{n,1}& \cdots &  U_{n,i}^{-1}U_{j,n}\Psi_{n,n}\\
\end{array}\right)
\end{align}
Then
\begin{align}
U[V_{i}^{T}U_{j}\circ\Psi]&=\left(\begin{array}
{ccc}
U_{1,1}& \cdots &U_{1,n}\\
\vdots & & \vdots\\
U_{n,1}& \cdots &U_{n,n}
\end{array}\right)
\left(\begin{array}{ccc}
U_{1,i}^{-1}U_{j,1}\Psi_{1,1} & \cdots & U_{1,i}^{-1}U_{j,n}\Psi_{1,n}\\
\vdots & & \vdots \\
U_{n,i}^{-1}U_{j,1}\Psi_{n,1}& \cdots &  U_{n,i}^{-1}U_{j,n}\Psi_{n,n}\\
\end{array}\right)\\ 
&=\left(\begin{array}{ccc}
\sum_{p=1}^{n}U_{1,p}U^{-1}_{p,i}U_{j,1}\psi_{p,1} & \cdots & \sum_{p=1}^{n}U_{1,p}U^{-1}_{p,i}U_{j,n}\psi_{p,n}\\
\vdots & & \vdots\\
\sum_{p=1}^{n}U_{n,p}U^{-1}_{p,i}U_{j,1}\psi_{p,1} & \cdots & \sum_{p=1}^{n}U_{n,p}U^{-1}_{p,i}U_{j,n}\psi_{p,n}
\end{array}\right)\\
U[V_{i}^{T}U_{j}\circ\Psi]U^{-1}&=\left(\begin{array}{ccc}
\sum_{p=1}^{n}U_{1,p}U^{-1}_{p,i}U_{j,1}\psi_{p,1} & \cdots & \sum_{p=1}^{n}U_{1,p}U^{-1}_{p,i}U_{j,n}\psi_{p,n}\\
\vdots & & \vdots\\
\sum_{p=1}^{n}U_{n,p}U^{-1}_{p,i}U_{j,1}\psi_{p,1} & \cdots & \sum_{p=1}^{n}U_{n,p}U^{-1}_{p,i}U_{j,n}\psi_{p,n}
\end{array}\right)\cdot\nonumber\\
&\qquad\qquad \left(
\begin{array}{ccc}
U_{1,1}^{-1} & \cdots &U_{1,n}^{-1}\\
\vdots & & \vdots\\
U_{n,1}^{-1}& \cdots &U_{n,n}^{-1}
\end{array}
\right)\\
&=
\left(
\begin{array}{ccc}
\sum_{q=1}^{n}\sum_{p=1}^{n}U_{1,p}U_{p,i}^{-1}U_{j,q}U^{-1}_{q,1}\Psi_{p,q} & \cdots &\sum_{q=1}^{n}\sum_{p=1}^{n}U_{1,p}U_{p,i}^{-1}U_{j,q}U^{-1}_{q,n}\Psi_{p,q}\\
\vdots & &\vdots \\
\sum_{q=1}^{n}\sum_{p=1}^{n}U_{n,p}U_{p,i}^{-1}U_{j,q}U^{-1}_{q,1}\Psi_{p,q} & \cdots & \sum_{q=1}^{n}\sum_{p=1}^{n}U_{n,p}U_{p,i}^{-1}U_{j,q}U^{-1}_{q,n}\Psi_{p,q}\\
\end{array}
\right)\\
&=
\left(
\begin{array}{ccc}
\sum_{p=1}^{n}U_{1,p}U_{p,i}^{-1}\sum_{q=1}^{n}U_{j,q}U^{-1}_{q,1}\Psi_{p,q} & \cdots &\sum_{p=1}^{n}U_{1,p}U_{p,i}^{-1}\sum_{q=1}^{n}U_{j,q}U^{-1}_{q,n}\Psi_{p,q}\\
\vdots & &\vdots \\
\sum_{p=1}^{n}U_{n,p}U_{p,i}^{-1}\sum_{q=1}^{n}U_{j,q}U^{-1}_{q,1}\Psi_{p,q} & \cdots & \sum_{p=1}^{n}U_{n,p}U_{p,i}^{-1}\sum_{q=1}^{n}U_{j,q}U^{-1}_{q,n}\Psi_{p,q}\\
\end{array}
\right)
\end{align}
So that
\begin{align}
[U[V_{i}^{T}U_{j}\circ\Psi(t)]U^{-1}]_{kl} &= \sum_{p=1}^{n}U_{k,p}U_{p,i}^{-1}\sum_{q=1}^{n}U_{j,q}U^{-1}_{q,l}\Psi_{p,q}(t)
\end{align}
as desired.
\end{document}